\pdfoutput=1

\documentclass[11pt]{article}

\usepackage[final]{acl}

\usepackage{times}
\usepackage{latexsym}
\usepackage{booktabs}

\usepackage[T1]{fontenc}

\usepackage[utf8]{inputenc}

\usepackage{microtype}
\usepackage{multirow}

\usepackage{inconsolata}

\usepackage{graphicx}

\usepackage{todonotes}

\title{LLMs can Perform Multi-Dimensional Analytic Writing Assessments: \\A Case Study of L2 Graduate-Level Academic English Writing}

\author{
Zhengxiang Wang\textsuperscript{\textdagger}\thanks{Zhengxiang Wang was a research assistant at the University of Saskatchewan for the research project that led to the creation of the corpus.} \hspace{.1cm} 
\textbf{Veronika Makarova}\textsuperscript{\textdaggerdbl} \hspace{.1cm} \textbf{Zhi Li}\textsuperscript{\textdaggerdbl} \hspace{.1cm} Jordan Kodner\textsuperscript{\textdagger} \hspace{.1cm} Owen Rambow\textsuperscript{\textdagger} \\
\textsuperscript{\textdagger}Department of Linguistics \& Institute for Advanced Computational Science, Stony Brook University \\ \textsuperscript{\textdaggerdbl}Department of Linguistics, University of Saskatchewan \\
\href{mailto:zhengxiang.wang@stonybrook.edu}{\texttt{zhengxiang.wang@stonybrook.edu}}
}

\begin{document}
\maketitle
\begin{abstract}

The paper explores the performance of LLMs in the context of multi-dimensional analytic writing assessments, i.e. their ability to provide both scores and comments based on multiple assessment criteria. Using a corpus of literature reviews written by L2 graduate students and assessed by human experts against 9 analytic criteria, we prompt several popular LLMs to perform the same task under various conditions. To evaluate the quality of feedback comments, we apply a novel feedback comment quality evaluation framework. This framework is interpretable, cost-efficient, scalable, and reproducible, compared to existing methods that rely on manual judgments. We find that LLMs can generate reasonably good and generally reliable multi-dimensional analytic assessments. We release our corpus and code\footnote{\url{https://github.com/jaaack-wang/multi-dimensional-analytic-writing-assessments}.} for reproducibility.

\end{abstract}

\section{Introduction\label{sec:introduction}}

\begin{figure}[]
    \centering
    \small
    \includegraphics[width=1\linewidth]{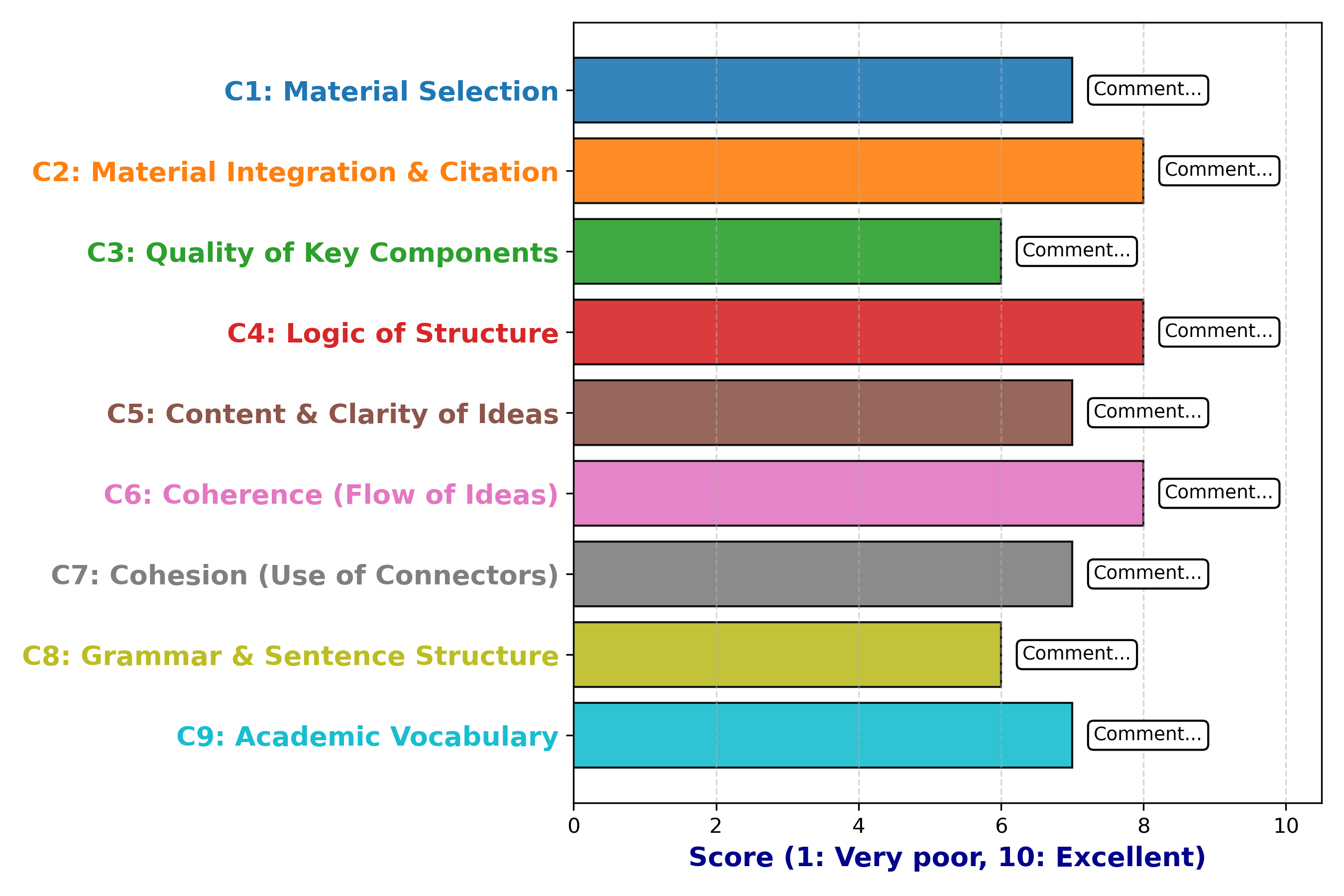}
    \caption{Multi-dimensional analytic assessments relevant to the corpus used in the study, where each assessment contains a score and a comment.}
    \label{fig:illustraion}
\end{figure}

Assessing the writing quality of essays manually is both time-consuming and labor-intensive. This task becomes even more demanding and challenging due to high cognitive load \citep{Cai2015}, when assessors have to assign scores and provide comments based on multi-dimensional analytic criteria, referred to here as \textit{multi-dimensional analytic assessments} (see Fig.~\ref{fig:illustraion} for an illustration). For evaluation of non-native language (L2) learners' writing, such precise and multi-dimensional assessments are highly valuable and desirable, but they are often not provided, due to the significant time, cost, and expertise required to produce them. This is also evidenced by the dearth of publicly available L2 writing corpora annotated with multi-dimensional analytic assessments \citep{banno-etal-2024-gpt}.

In recent years, large language models (LLMs) have emerged as promising tools for self-regulated writing assessments among L2 learners. A growing number of studies \citep[][\textit{i.a.}]{chiang-lee-2023-large, MIZUMOTO2023100050, han-etal-2024-llm, yancey-etal-2023-rating} have indicated the general usefulness of LLMs for automated writing assessments. Given their increasing use for this task, the following question remains understudied: \textit{can LLMs provide reasonably good multi-dimensional analytic writing assessments?} We use the phrase ``reasonably good'' intentionally, given the open-ended nature of the task, particularly generating essay-level feedback comments.\footnote{Strictly speaking, there are two types of feedback: quantitative feedback (scores) and qualitative feedback (comments). We use ``feedback comments'' to refer to qualitative feedback.}

To address this question, we utilize an English-language corpus of literature reviews written by L2 graduate students and assessed by human experts on 9 analytic assessment criteria. We prompt various popular LLMs to assess the corpus using the same criteria under various conditions, and we examine the quality of their generated assessments compared to human-generated assessments.

\smallskip
\noindent Our study makes three primary contributions: 

\begin{enumerate}
    \item We provide empirical evidence that LLMs can generate reasonably good and generally reliable multi-dimensional analytic writing assessments. This is the primary goal of this study; we do not argue in favor of a specific LLM, nor do we advocate replacing humans with LLMs for this task.
    
    \item We release a corpus of L2 English graduate-level literature reviews, annotated with multi-dimensional analytic assessments, which will facilitate future studies.
    
    \item We propose and validate a novel LLM-based framework, ProEval, for evaluating the quality of feedback comments. ProEval is time- and cost-efficient, scalable, and reproducible, compared to manual judgments. It is also interpretable and fine-grained, compared to direct quality ratings.

\end{enumerate}

\section{Related Work\label{sec:relatedWork}}

\paragraph{Automated Writing Evaluation (AWE)} We use AWE to include both automated essay scoring (AES)\footnote{AES is sometimes conflated with AWE in the literature \citep{Hockly2019}. We distinguish them.} and feedback comment generation \citep{Shermis2013-ja}. AWE systems have existed since the 1960s \citep{page1966imminence} and have evolved over time with a predominant focus on AES \citep{Ke2019, Hussein2019, Zhang2020, Uto2021, Lagakis2021}. Modern AWE systems use deep neural networks for scoring \citep{taghipour-ng-2016-neural, alikaniotis-etal-2016-automatic, dong-etal-2017-attention, rodriguez2019languagemodelsautomatedessay, yang-etal-2020-enhancing, xie-etal-2022-automated} and feedback comment generation \citep{nagata-2019-toward, han-etal-2019-level, babakov-etal-2023-error}. The latter task typically focuses on sentence-level grammatical error identification and correction \citep{behzad-etal-2024-leaf}. Existing non-LLM AWE systems mainly provide holistic assessment, with some specialized systems offering uni-dimensional analytic assessment based on a specific dimension of writing quality \citep{Ke2019, jong2023reviewfeedbackautomatedessay, banno-etal-2024-gpt}.

\paragraph{LLMs used for AWE} Unlike prior AWE systems, LLMs can be prompted in natural language to jointly score and comment on a given essay. A growing body of research has explored the use of LLMs for assessing L2 writing. For AES, LLMs have been examined for holistic scoring \citep{MIZUMOTO2023100050, yancey-etal-2023-rating, Wang2024}, discourse coherence scoring \citep{naismith-etal-2023-automated}, and multi-dimensional analytic scoring \citep{Yavuz2024, banno-etal-2024-gpt}. For feedback comment generation, LLMs have been studied for generating corrective comments \citep{MIZUMOTO2024100116, song-etal-2024-gee}, holistic comments \citep{behzad-etal-2024-assessing, behzad-etal-2024-leaf}, and multi-dimensional analytic comments \citep{Guo2024-eg, behzad-etal-2024-assessing, han-etal-2024-llm}. \citet{stahl-etal-2024-exploring} is the only study we know of which explores LLMs jointly performing scoring and feedback comment generation, but holistically. Moreover, the ASAP\footnote{\url{https://www.kaggle.com/competitions/asap-aes}} corpus they use contains short essays by native speakers from Grade 7 to Grade 10 and has no human reference comments.  

\paragraph{Related Corpora} Major L2 writing corpora include TOEFL11 \citep{TOEFL11}, which contains scored essays from TOEFL tests, and CLC-FCE \citep{yannakoudakis-etal-2011-new}, which includes error-annotated short texts in response to exam prompts. Other notable corpora are derived from online language learning platforms, such as EFCAMDAT \citep{vanRooy_2015}, Write \& Improve \citep{Helen2018}, and LEAF \citep{behzad-etal-2024-leaf}, focusing on scoring, grammatical error correction, and personalized feedback, respectively. We are not aware of any publicly available corpora annotated with multi-dimensional analytic scores and comments jointly.

\section{Corpus\label{sec:corpus}}

\paragraph{Overview} Our corpus consists of 141 literature reviews written in English by 51 L2 graduate students, with an average word count of 1321 (930 excluding references). The reviews cover five broad topics from the humanities and social sciences, chosen to minimize the need for specialized disciplinary knowledge: (1) the social consequences of legalized cannabis, (2) the Canadian linguistic landscape, (3) online learning, (4) lessons from the COVID-19 pandemic, and (5) pacifism. Essays on topics 1, 3, and 5 were written individually, while those on topics 2 and 4 were completed collaboratively by 2-4 authors.

The corpus is a result of a large research project conducted at the University of Saskatchewan, a Canadian public research university, in 2021 with an aim to examine the developmental trajectory of literature review writing skills among L2 graduate students. The project involved three rounds of a 5-unit online tutorial series conducted over the course of 2021, with each round lasting 13 weeks (see Appendix~\ref{app:corpus} for details). Participation was voluntary, with 31 participants completing all five writing tasks across all rounds, and 20 further students completing at least one task before withdrawing.

\paragraph{Our Previous Studies}
The corpus has been used in our previous studies \citep{li2023assessment, li2023developing, makarova2024can}, although it has not been made public until now. These three studies only use a subset of the corpus, namely essays written individually or those based on topics 1, 3, and 5. 

Among these studies, \citet{li2023assessment, li2023developing} focus on individual writing development without examining feedback comments, placing their work within English for Academic Purposes rather than AWE. While \citet{makarova2024can} explore ChatGPT’s ability to assess L2 academic writing, they only compare model output to averaged human scores and aggregated comments, lacking criterion-level analysis. Their analysis is limited to surface features (e.g., word count, type-token ratio, comment length) and does not consider different prompting conditions. In contrast, this study offers a broader evaluation using the full corpus, distinct methodologies, and a more fine-grained analysis, with no substantial overlap with our prior work.

\paragraph{Essay Authors} The corpus authors comprise a diverse group of L2 learners, representing a wide range of first languages and enrolled in graduate programs across various disciplines at multiple Canadian universities. Their English proficiency ranged from upper-intermediate to advanced, with an average score equivalent to IELTS band score\footnote{\url{https://ielts.org/take-a-test/your-results/ielts-scoring-in-detail}.} 7 based on conversions from various standardized English language tests. Scores varied from IELTS 6.5 to 8.5, with a standard deviation of 0.55.

To support their writing, authors received a curated bibliography for each writing task, designed to facilitate literature review writing while reducing the burden of bibliographic searches. Before submitting their final drafts for expert assessments, they participated in peer review (for topics 1, 3, and 5) or group collaboration (for topics 2 and 4). These two measures were intended to enhance the overall quality of the submitted essays.

\paragraph{Human Assessments} Most essays in the corpus were assessed by three (94.3\%) or two (5.0\%) independent human experts. As illustrated in Fig.~\ref{fig:illustraion}, the assessments consist of scores on a 10-point scale and comments based on 9 analytic assessment criteria. While scores were required, comments were optional for the assessors. Six assessors with professional experience in English language teaching assessed at different stages of the research project. Table~\ref{tab:feedbackRate} provides basic information about them.

The 9 assessment criteria include: (C1) material selection; (C2) material integration and citation; (C3) quality of key components; (C4) logic of structure; (C5) content and clarity of ideas; (C6) coherence (flow of ideas) ; (C7) cohesion (use of connectors); (C8) grammar and sentence structure; and (C9) academic vocabulary. Comparatively, criteria C2, C8, and C9 are more technical and objective, since there are clearer rules and conventions governing proper citation practices, grammatical correctness, and appropriate academic word usage. In contrast, other criteria require more interpretive judgment, making them relatively more subjective in nature. See Table~\ref{tab:assessmentCriteria} in Appendix~\ref{app:criteria} for details about these criteria.

\paragraph{Assessment Quality}  The 31 students who completed all writing tasks evaluated the quality of human assessments on a 4-point scale in an anonymous final project survey. Based on the 30 submitted survey responses, all participants agreed that the assessments were at least ``useful'' (rating = 3), with 24 participants (80\%) rating them as ``very useful'' (rating = 4). 

\paragraph{Data Contamination} Since the corpus was created prior to the release of ChatGPT and has never been made public, it contains no LLM-generated contents and is free from the risk of data contamination \citep{jacovi-etal-2023-stop, sainz-etal-2023-nlp}, making it an ideal resource for LLM evaluation.

\begin{table}[]
    \centering
    \small
    \begin{tabular}{lllllll}
    \toprule
    Code & Role & Rounds & Topics & \# Essays \\
    
    \midrule
    A & Graduate RA & 1 & 1-5  &  27 \\
    B & Graduate RA &  1-3 & 1-5 &   141  \\
    C & Faculty Member & 1-3 & 1, 2, 5 &  93 \\
    D & Faculty Member & 1 & 2 &  4 \\
    E & Faculty Member & 1-3 & 3, 4 &  43 \\
    F & Graduate RA & 2, 3 & 1-5 & 106  \\
    \bottomrule
    \end{tabular}
    
    \caption{Anonymized information for the six assessors (A–F). The columns ``Rounds'' and ``Topics'' indicate the specific rounds and writing topics they participated in. Assessors C and E never co-assessed together.}
    \label{tab:feedbackRate}
\end{table}

\begin{figure*}[ht]
    \centering
    \small
    \includegraphics[width=1\linewidth]{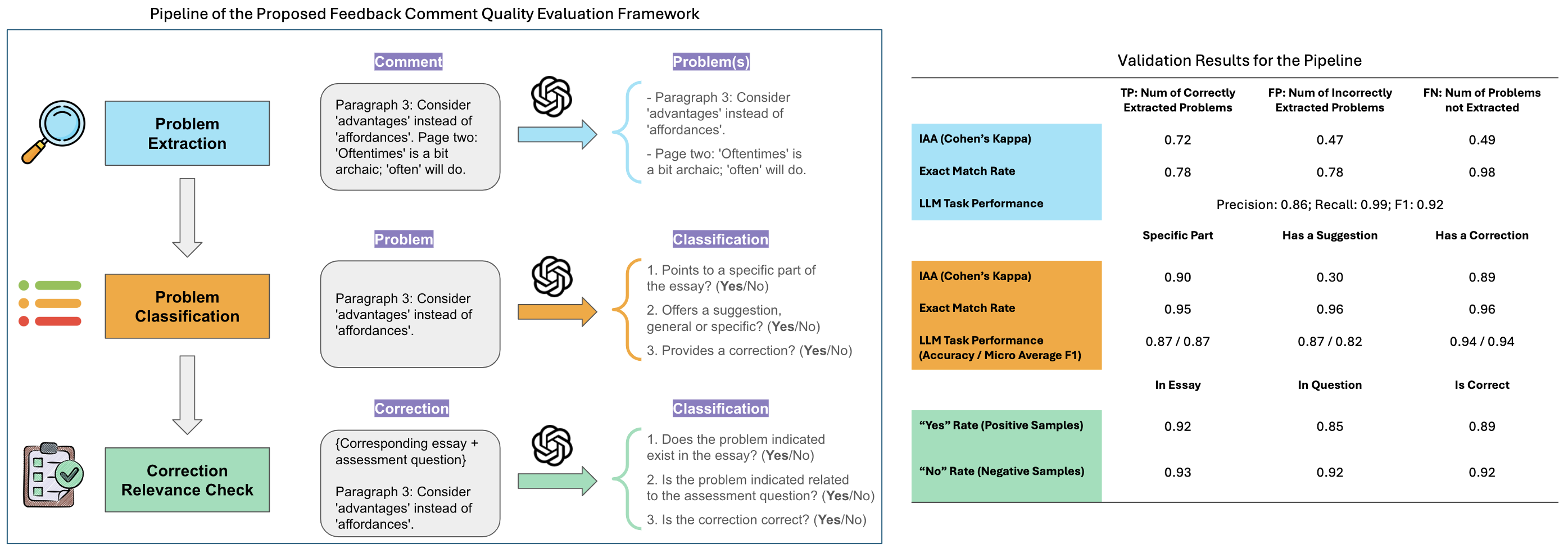}
    \caption{\textit{Left}: Pipeline of the proposed feedback comment quality evaluation framework. The input and output for each step of the pipeline are illustrated using a human-generated comment on the use of academic vocabulary, with related tasks performed by an LLM. Answers to the 6  classification questions from the last two steps are highlighted in bold. \textit{Right}: Validation results for the pipeline, where IAA (inter-annotator agreement) and exact match rate are measured between raw annotations by two annotators. See Appendix~\ref{app:framework} for further details.}
    \label{fig:novelCommentEvalFramework}
\end{figure*}


\section{ProEval: A Novel Feedback Comment Quality Evaluation Framework\label{sec:evalOfComments}}

A common approach to evaluating feedback comment quality for an
essay uses manual judgments (e.g., rating on a Likert scale), since generating essay-level feedback is an open-ended task. However, this approach is expensive, time-consuming, not scalable, and may not always be reproducible. 

For L2-related feedback comments, common criteria for assessing comment quality include specificity, relevance, helpfulness \citep{han-etal-2024-llm, stahl-etal-2024-exploring, behzad-etal-2024-assessing, behzad-etal-2024-leaf}, and the ability to identify writing problems \citep{stahl-etal-2024-exploring, behzad-etal-2024-assessing, behzad-etal-2024-leaf}. These criteria reflect a common and practical need of L2 learners to be shown specific problems in their essays and how to correct them to improve their writing quality.

\subsection{The Framework}

To address the issues of manual judgment, we propose \textbf{ProEval}, an automatic \textbf{pro}blem-focused \textbf{eval}uation framework that \textit{evaluates the quality of a feedback comment in terms of its ability to effectively identify relevant writing problems within the assessed essay.} As illustrated in Fig.~\ref{fig:novelCommentEvalFramework} (left), the framework utilizes LLMs to extract problems identified in feedback comments and to characterize their specificity and potential helpfulness. Rather than prompting an LLM to directly rate comment quality, which raises concerns about the reliability of LLM-based evaluators \cite{doddapaneni-etal-2024-finding}, our approach prioritizes transparency and interpretability by grounding evaluation in concrete, observable features.

More concretely, ProEval consists of the following three steps, with the first two steps automated by \textsc{GPT-4o-2024-11-20} \cite{openai2024gpt4ocard} and the last step by \textsc{GPT-4-Turbo-2024-04-09} \cite{openai2023gpt4}. See Appendix~\ref{app:framework} for additional details and explanations.

\paragraph{Problem Extraction} We start out by extracting any writing problems stated or implied in assessment comments, along with any relevant contextual information for each problem, such as further explanations, suggestions for improvement, concrete corrections, or clarifying questions. We define a problem as any writing-related issue that affects the quality of the writing, such as citation errors, logical flaws, or grammatical mistakes.

\paragraph{Problem Classification} The extracted problems are further characterized along three dimensions: whether an extracted problem \textbf{(1)} points to a specific part of the essay, \textbf{(2)} includes any form of suggestion (general or specific), and \textbf{(3)} provides a concrete correction that can be directly applied to fix an identified problem. These classifications offer a quantifiable way to assess the \textit{specificity} and \textit{potential helpfulness} of related comments.

\paragraph{Correction Relevance Check} We perform a sanity check to determine whether the proposed correction (and thus the comment) is in fact relevant to the original essay. The Correction Relevance Check also contains three binary classification questions for a more nuanced relevance analysis: \textbf{(1)} does the problem indicated in the correction exist in the essay? \textbf{(2)} is the indicated problem related to the given assessment question? and \textbf{(3)} is the correction correct? 

The results show that both human- and LLM-provided corrections are highly relevant, with answers to those three questions being ``Yes'' typically above 90\% time (see Table~\ref{tab:overallRelevanceCheckResults} in Appendix~\ref{app:relevanceCheck}). We thus focus on the Problem Classification results in the next two sections.  


\subsection{Validations of the Framework}


The basic idea of ProEval is to break down a complex and inherently subjective evaluation task into multi-level subtasks that are easy for humans to verify and well-suited for LLMs to perform. To validate that, the first author and a paid graduate student in Linguistics (native speaker) first annotated some held-out samples for training and developing the annotation guidelines. Each then independently annotated at least another 200 samples containing human- and LLM-generated comments or problems for Problem Extraction and Problem Classification. Afterward, they met to resolve disagreements before the inter-annotator agreement (IAA) was calculated.

We measure IAA using Cohen's Kappa. As is known \cite{Feinstein1990}, Cohen's Kappa can provide misleading values with highly imbalanced class distributions. We therefore also provide exact match rates which have not been corrected for random agreement. Fig.~\ref{fig:novelCommentEvalFramework} (right) shows that the IAA is typically high. When the Cohen's Kappa is low due to class imbalance (i.e., problems being incorrectly or not extracted is uncommon or rare and nearly all extracted problems contain a suggestion), the exact match rates are high. LLM task performance, evaluated based on the resolved annotations, is also notably high (e.g., 0.92 F1 for Problem Extraction and at least 87\% accuracy for the classification tasks in Problem Classification).

We automatically evaluate LLM performance on the Correction Relevance Check by assuming that human-identified corrections are generally relevant. Specifically, we assess whether the LLM classifies these corrections as mostly \textit{relevant} when presented with their corresponding essays and assessment questions (positive samples), and as mostly \textit{irrelevant} when paired with random essays and questions (negative samples). As shown in Fig.~\ref{fig:novelCommentEvalFramework} (right), our results confirm this expectation.

\section{Experiments\label{sec:experiments}}

This sections describes and presents the main experiments conducted and the results obtained.

\subsection{LLM Prompting\label{sec:prompting}}

\paragraph{List of LLMs} We evaluate variants of three popular LLMs: \textsc{gpt-4o-2024-08-06} (GPT-4o, \citealp{openai2024gpt4ocard}), \textsc{Gemini-1.5-flash} (Gemini-1.5, \citealp{geminiteam2024gemini15}), and \textsc{Llama-3 70B-Instruct} (Llama-3, \citealp{grattafiori2024llama3herdmodels}).

\paragraph{Default Prompt Setting} All prompts contain a system prompt, an input essay, and an assessment instruction. There are four default conditions. \textbf{(1)} The system prompt contains not only essential background information, such as writing topic, but also helpful information regarding the L2 nature of the input essay, year of writing, the same general assessment guidance used by human assessors. \textbf{(2)} The input essay always includes references. \textbf{(3)} LLMs are instructed to produce a score before an optional comment for each assessment question \textbf{(4)} via greedy decoding, i.e., with temperature set to 0. Conditions 1-3 are used to maximize the alignment between human and LLM assessment conditions.

\paragraph{Interaction Modes} We consider three possible user-LLM interaction modes, depending on how the 9 assessment questions are presented. In Interaction Mode 1 (IM 1), all questions are prompted at once in a single-turn conversation, where all LLM assessments are generated in a single response. In Interaction Mode 2 (IM 2), the questions are asked one at a time, with an LLM generating answers to each question in corresponding turns in a multi-turn conversation. In Interaction Mode 3 (IM 3), however, the assessment questions are provided independently of one another in 9 separate prompts to elicit 9 separate outputs from an LLM.

\subsection{Baselines}

Given the open-ended nature of the task, we compare raw assessments produced across individual assessors to understand the assessment patterns and behaviors of humans and LLMs. For a more robust statistical analysis, we only consider raw assessments made by assessors B, C, and F, since the essays they each assessed and co-assessed both cover at least half of the corpus (at least 78 essays between assessors C and F). See Table~\ref{tab:number_CoAssessedEssays} in Appendix~\ref{app:numberCoAssessed} for exact numbers of essays all assessor pairs (including LLM assessors) co-assessed.

\subsection{Evaluation of Scores\label{sec:evalOfScores}}

\paragraph{Quadratic Weighted Kappa (QWK)} This is a metric for rating inter-rater agreement. It ranges from 0 (random agreement) to 1 (perfect agreement), though it can be negative when agreement is worse than chance. QWK places higher penalties for larger score mismatches, but can yield misleadingly high or low values due to chance correction when the distribution of scores is highly skewed \citep{yannakoudakis-cummins-2015-evaluating}.

\paragraph{Adjacent Agreement Rate (AAR)} AAR measures the percentage of scores (from two raters) that lie within a specified threshold $k$ of one another. When $k=0$, it assesses exact matches. For this study, we set $k=1$ (AAR1), meaning raters' scores are treated as matching or equivalent as long as they differ by no greater than 1.

We use AAR1 in addition to QWK to account for the limitation of QWK's chance correction, as we observe that both human- and LLM-assigned scores are highly biased toward the respective means. AAR1 also helps address observed scoring inconsistency issues (often by 1 point) by humans. See Appendix~\ref{app:scores} for more details and discussions.

\begin{figure}
    \centering
    \includegraphics[width=1\linewidth]{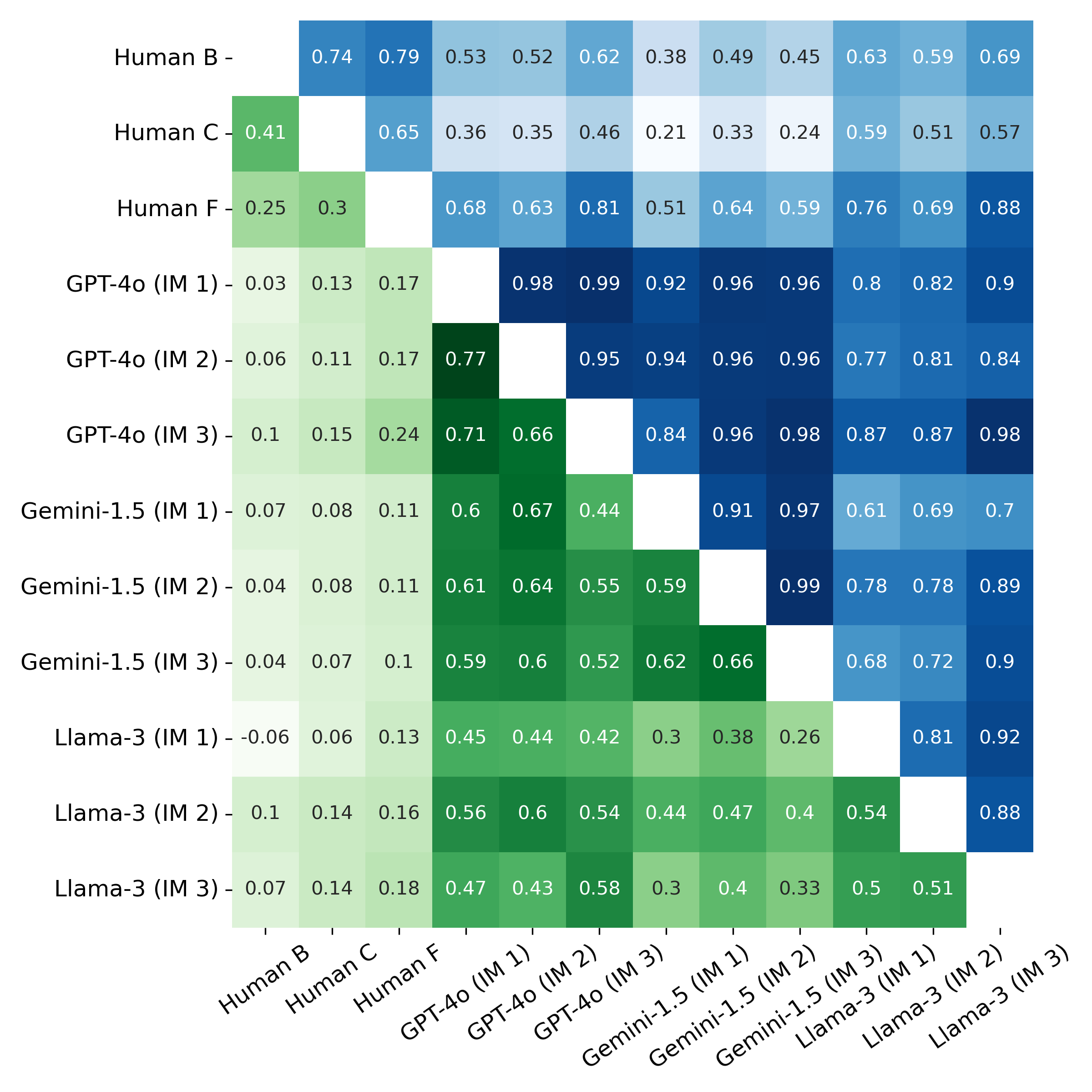}
    \caption{Heatmaps of overall QWK (bottom, green) and AAR1 (top, blue) among assessors. Darker shades indicate a higher degree of agreement.}    \label{fig:overall_agreement_heatmap}
\end{figure}

\begin{figure}
    \centering
    \small
    \includegraphics[width=0.95\linewidth]{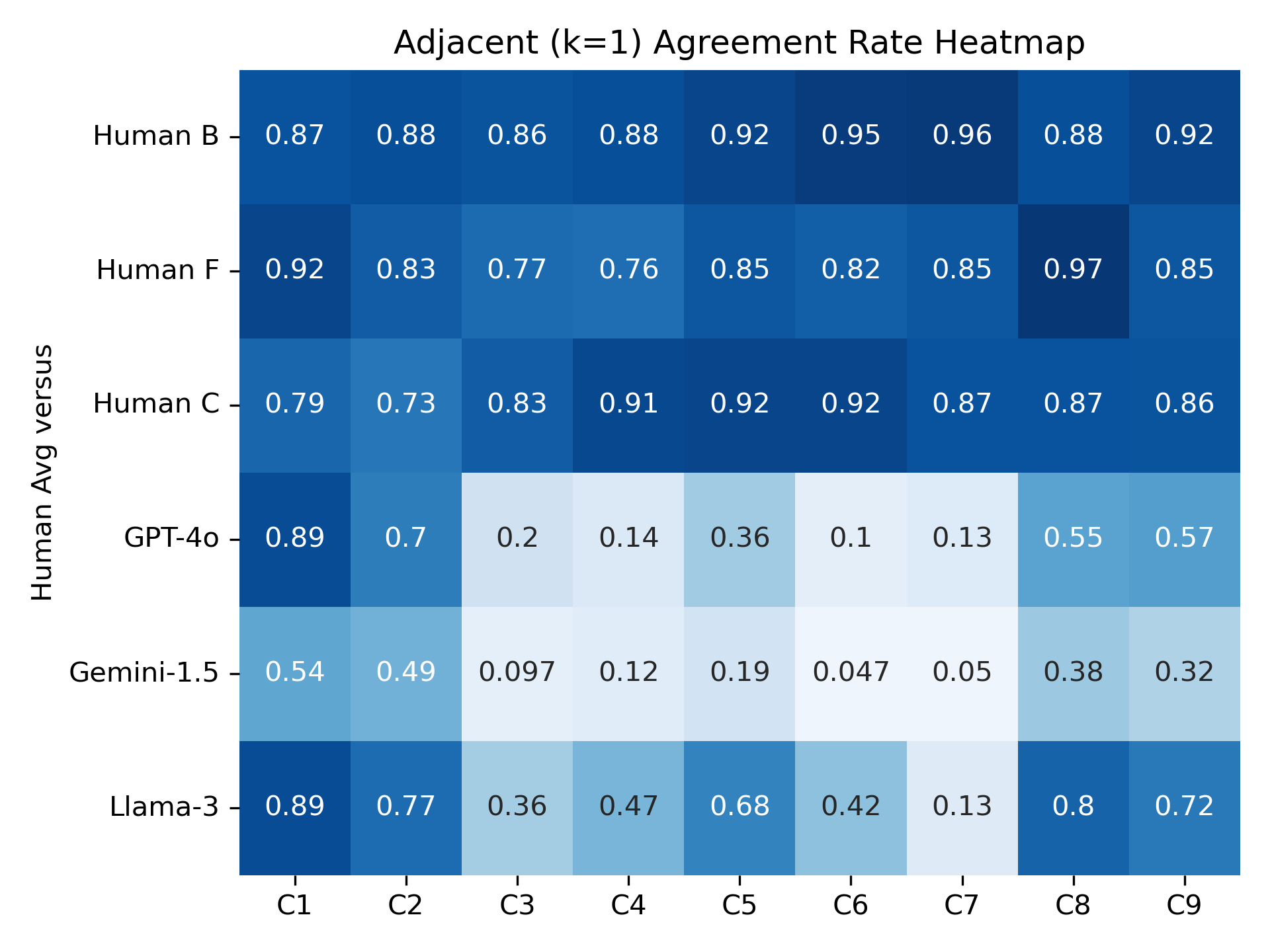}
    \caption{Criterion-level AAR1 between average human scores (``Human Avg'') and human or LLM assessors. See Appendix~\ref{app:scores} for full results for QWK and AAR1.}
    \label{fig:criterion-level_agreement_heatmap}
\end{figure}

\subsection{Results}

We compare human- and LLM-generated assessments in terms of scores, comments, and the interaction between scores and comments.

\subsubsection{Scores\label{sec:scores}}

Fig.~\ref{fig:overall_agreement_heatmap} illustrates the overall scoring agreement between all pairs of assessors. 

\paragraph{Humans score more like humans and LLMs score more like LLMs.} More concretely, human-human QWK and AAR1 are almost always higher than the corresponding human-LLM agreement. Similarly, LLM-LLM agreement exceeds human-LLM agreement in virtually all cases, with a much larger margin, suggesting that LLMs may resemble each other in scoring more closely than humans resemble each other. This may be attributed to the substantial overlap in LLM training data, in contrast to the broader variability in human linguistic experiences, which contributes to greater divergence in human scoring patterns. Criterion-level agreement between human/LLM assessors shows similar patterns, as shown in Fig.~\ref{fig:criterion-level_agreement_heatmap}.

\paragraph{LLMs can score approximately like humans.} The best human-LLM AAR1 for the three LLMs ranges from 0.59 to 0.88, with all LLMs achieving an AAR1 above 0.5 with assessor F (Fig.~\ref{fig:overall_agreement_heatmap}). Moreover, the AAR1 scores between GPT-4o and assessor B and between Llama-3 and assessors B and C are always greater than 0.5. \textit{Overall, it shows that LLMs can generate sensible or reasonably good scores, often differing by no more than 1 point from the corresponding human-generated scores.}

\paragraph{Human-LLM agreement tends to be higher when LLMs respond to each assessment criterion separately under IM 3.} This is particularly true compared to when LLMs respond to all criteria at once under IM 1, since IM 3 exhibits a generally higher agreement level (Fig.~\ref{fig:overall_agreement_heatmap}). This result may imply that, while human assessors score the 9 assessment criteria sequentially, they effectively make independent scoring decisions based on the specifics of each assessment question. 

That said, the effect of interaction modes is overall limited, given the fairly close scores (i.e., high QWK/AAR1) assigned across them for each LLM. Therefore, we average human-LLM agreement for each LLM across the three interaction modes to obtain human-LLM agreement in Fig.~\ref{fig:criterion-level_agreement_heatmap}.

\paragraph{The degree of human-LLM agreement varies across assessment criteria.} For example, Fig.~\ref{fig:criterion-level_agreement_heatmap} shows that LLM-assigned scores are relatively closer to human-assigned scores on assessment criteria C1 (material selection), C2 (material integration and citation), C8 (grammar and sentence structure), and C9 (academic vocabulary) than the other criteria. Among criteria C3-C7, LLMs and humans agree rather poorly on C7 (use of connectors), with LLMs consistently assigning scores more than 1 point away from human-assigned ones.

\begin{table}[]
    \centering
    \setlength{\tabcolsep}{5pt}
    \scriptsize

\begin{tabular}{llllll}
\toprule
                  &    \multicolumn{2}{l}{Avg Comment} & \multicolumn{2}{l}{Avg Problem} & Avg Corr \\
         Assessor &    Rate &             Len &    Rate &             Num & Score - Cmt \\
\midrule
          Human B &  0.24 &  104{\tiny±85} &  0.97 & 3.8{\tiny±3.5} & \textbf{-0.20} / -0.17 \\
          Human C &  1.00 &   62{\tiny±85} &  0.56 & 1.3{\tiny±1.8} & -0.40 /     \textbf{-0.46} \\
          Human F &  0.90 &   47{\tiny±58} &  0.63 & 1.3{\tiny±1.6} & -0.37 /     \textbf{-0.47} \\ \midrule
          
          GPT-4o (IM 1) & 1.00 &   65{\tiny±14} & 1.00 & 2.1{\tiny±0.9} & -0.11 /      \textbf{-0.48} \\
          Gemini-1.5 (IM 1) & 1.00 &   97{\tiny±33} & 1.00 & 2.4{\tiny±1.00} & -0.05 /      \textbf{-0.46} \\
          Llama-3 (IM 1) & 1.00 &   68{\tiny±20} & 1.00 & 2.2{\tiny±0.8} &  0.01 /     \textbf{-0.27} \\ \midrule

          GPT-4o (IM 2) & 1.00 &  347{\tiny±46} & 1.00 & 5.0{\tiny±1.2} & -0.37 /     \textbf{-0.38} \\
          Gemini-1.5 (IM 2) & 1.00 & 477{\tiny±698} & 1.00 & 5.9{\tiny±2.7} & -0.29 /      \textbf{-0.56} \\
          Llama-3 (IM 2) & 1.00 & 370{\tiny±112} & 1.00 & 6.6{\tiny±2.8} & -0.04 /     \textbf{-0.42} \\ \midrule
          
          GPT-4o (IM 3) & 1.00 &  381{\tiny±65} & 1.00 & 6.1{\tiny±2.0} & -0.34 /      \textbf{-0.51} \\
          Gemini-1.5 (IM 3) & 1.00 & 571{\tiny±182} & 1.00 & 8.2{\tiny±3.3} & -0.21 /      \textbf{-0.48} \\
          Llama-3 (IM 3) & 1.00 &  399{\tiny±67} & 1.00 & 6.4{\tiny±2.3} & -0.04 /    \textbf{-0.23} \\
\bottomrule
\end{tabular}
    
    \caption{Overall statistics of feedback comments generated by human and LLM assessors. The last column shows the Spearman Rank correlations measured between scores and related comments (length / number of identified problems). Stronger negative correlations (smaller numbers) in each number pair are in bold.}
    \label{tab:overallCommentStats}
\end{table}



\begin{figure*}
    \centering
    \small
    \includegraphics[width=1\linewidth]{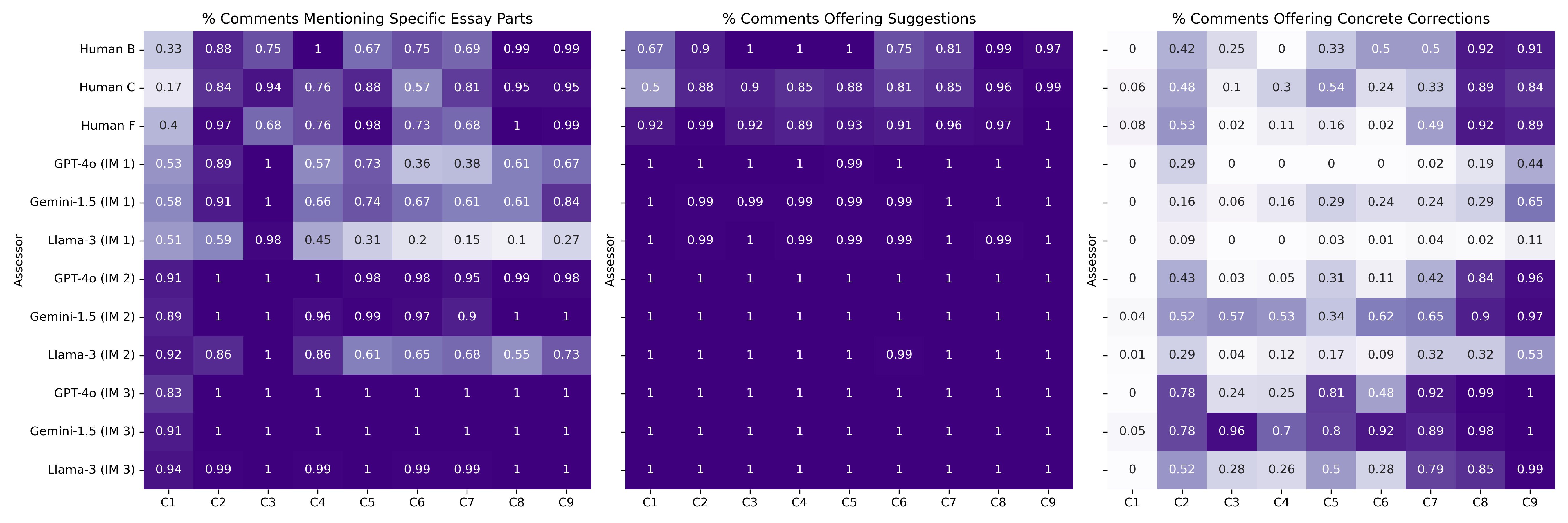}
    \caption{Percentage of comments identifying a problem that mentions a specific essay part (left), offers a comment (middle), and offers a concrete correction (right) across assessment criteria by different assessors.}
    \label{fig:criterion_level_problemChars}
\end{figure*}

\subsubsection{Comments}

Table~\ref{tab:overallCommentStats} shows the percentage of time an assessor provided a comment, and when they did, the average length of these comments, the percentage of comments identifying a problem, and the average number of problems identified in each comment. 

\paragraph{LLMs always provide comments and identify problems, but humans do not.} This is an apparent advantage of LLMs since, unlike humans, they do not experience practical constraints like mental fatigue and limited time for writing comments. While humans show different tendencies in comment writing, they tend to write more comments and/or identify more problems (with longer comments) on criteria that are technical and objective, including C2, C8, and C9, also mentioned in the end of Section~\ref{sec:scores}. See Appendix~\ref{app:comments} for details.

\paragraph{Interacting with LLMs one question at a time leads to more elaborate, specific, and helpful comments.} LLM comments are much longer and identify more problems in IM 2 and IM 3 than in IM 1 (see Table~\ref{tab:overallCommentStats}). Additionally, Fig.~\ref{fig:criterion_level_problemChars} shows that comments generated in IM 1 are also less likely to refer to a specific essay part and offer a concrete correction than those generated in IM 2 and IM 3 or human-generated comments. This suggests that IM 2 and IM 3 provide higher levels of elaboration than IM 1. Furthermore, IM 3 produces more corrections than both IM 2 and humans across all assessment criteria, except C1, for which a correction is unlikely since it is about evaluating the relevance of cited references. In other words, \textit{LLMs can be more elaborate, specific, and potentially helpful than humans in their comments.}

\paragraph{LLMs can be  more specific than humans on assessing subjective criteria.} While humans and LLMs (in IM 3) are comparably likely to include a correction in their comments for objective criteria C2, C8, and C9, LLMs' comments (in IM 3) tend to offer more corrections on other subjective criteria (e.g., C3: quality of key components, C4: logic of structure etc.), except for C1 (see above). This aligns with the observation that humans tend to comment more on objective criteria, since commenting on subjective criteria requires more explanations and can thus be more demanding to do.

\subsubsection{Score-Comment Interaction} 

Since lower scores reflect a perception of more writing problems, an assessor typically needs to provide a more extensive feedback comment to both cover the identified problems and justify their low scores. We highlight this score-comment interaction by measuring the correlations between scores and the token counts of or the numbers of identified problems in the related comments. 

As expected, the last column in Table~\ref{tab:overallCommentStats} shows strongly negative score-comment correlations across both human- and LLM-generated assessments. The fact that these negative correlations are generally much stronger when measured with the number of identified problems suggests that it is a more fine-grained metric than comment length and also indicates the usefulness of our framework (ProEval) proposed in Section~\ref{sec:evalOfComments}. See Fig.~\ref{fig:score-comment-interaction} in Appendix~\ref{app:score-comment-interaction} for full results of the correlations.

\subsection{Summary}

We show that LLMs can generate sensible scores, typically within 1 point of human-generated ones on a 10-point scale, and feedback comments that identify more writing problems than human assessors that are specific, and potentially helpful. This is particularly true when LLMs are prompted in IM 3 where each assessment question is asked independently of each other. Moreover, like humans, LLMs also generate assessments that exhibit an expected and negative score-comment correlation, justifying the validity of their assessments. \textit{Overall, these results highlight that LLMs can generate reasonably good multi-dimensional analytic assessments.} 


\section{Further Analyses\label{sec:furtherAnalyses}} 

This section reexamines the assumption underlying our proposed feedback comment quality evaluation framework, i.e., ProEval, and evaluates the reliability of LLM-generated assessments.

\begin{table}[]
    \centering
    \scriptsize

\begin{tabular}{llrrr}
\toprule
        &             &  \#Problems &  \#Specific &  \#Corrections \\
Condition &  &            &                 &               \\
\midrule
Humans & Specificity &       0.57 &            0.66 &          0.63 \\
        & Helpfulness      &       0.65 &            0.70 &          0.62 \\
LLMs & Specificity      &       0.62 &            0.80 &          0.61 \\
        & Helpfulness      &       0.64 &            0.77 &          0.58 \\ \midrule
        
C6 & Specificity &       0.68 &            0.78 &          0.51 \\
        & Helpfulness &       0.72 &            0.74 &          0.48 \\

C9 & Specificity &       0.59 &            0.79 &          0.77 \\
        & Helpfulness &       0.64 &            0.76 &          0.74 \\ \midrule

IM 1 & Specificity &       -0.02 &            0.63 &          0.43 \\
        & Helpfulness &      -0.03 &            0.50 &          0.44 \\

IM 2 & Specificity &       -0.02 &            0.63 &          0.43 \\
        & Helpfulness &     0.09 &            0.48 &          0.38 \\

IM 3 & Specificity &     0.22 &            0.33 &          0.31 \\
        & Helpfulness &     0.23 &            0.30 &          0.24 \\

\bottomrule
\end{tabular}

    \caption{Spearman Rank correlations between the specificity and helpfulness scores and the number of different types of problems identified by our framework under different conditions. Corrections with number of problems making a suggestion are omitted as they are nearly identical to those with ``\#Problems.''}
    \label{tab:specificityHelpfulnessCorr}
\end{table}

\subsection{Re-examining Our Assumption about Feedback Comment Quality} 

ProEval assumes that the quality of a feedback comment is related to how well it identifies relevant writing problems of an assessed essay. The framework extracts and characterizes problems of assessed essays identified in comments to evaluate the specificity and helpfulness of these comments.

To assess this assumption, we adopt an LLM-as-a-judge approach \cite{zheng2023judgingllmasajudge}, prompting \textsc{openai-o1-mini-2024-09-12} (o1-mini, \citealp{openai2024openaio1card}) to directly assess the specificity and helpfulness of a feedback comment, given the corresponding essay and assessment question on a 10-point scale. We do not define specificity and helpfulness to avoid injecting biases and choose all comments, generated by humans and LLMs, from one subjective criterion (C6: coherence or flow of ideas) and one objective criterion (C9: academic vocabulary) to balance our examination. We then calculate the average Spearman rank correlations between these two scores produced by o1-mini and the number of different types of problems identified by ProEval under varying conditions. 

The results in Table~\ref{tab:specificityHelpfulnessCorr} shows that the characteristics extractable from applying ProEval correlate very well with the o1-mini-assigned specificity and helpfulness scores. In particular, the number of problems that mention specific essay parts and offer corrections appears to be overall stronger signals of specificity and helpfulness than the mere number of problems, which shows negligible correlations for comments from IM 1 or IM 2. This shows the potential of ProEval in providing a more fine-grained and interpretable measurement of specificity and helpfulness levels of comments.

\begin{table}[]
    \centering
    \scriptsize
    
\begin{tabular}{p{2.65cm}|p{1.5cm}p{1.75cm}}
\toprule            
                   &  Scores &  Comments \\
\midrule
                 GPT-4o-May & 0.82 / 0.98 &  0.21 / 0.39 / 0.70  \\
           SP Simplification & 0.78 / 0.98 &  0.24 /    0.43 /       0.72 \\
                Exclusion of References & 0.69 / 0.95 &  0.26 /    0.44 /       0.73 \\
               Comment First & 0.75 / 0.96 &  0.19 /    0.32 /       0.58 \\
                  Temperature=1, run\#1 & 0.73 / 0.96 &  0.10 /    0.30 /       0.67 \\
                  Temperature=1, run\#2 & 0.79 / 0.98 &  0.10 /    0.31 /       0.67 \\ \midrule \midrule
             GPT-4o-May (IM 2) & 0.81 / 0.99 &  0.15 /    0.29 /       0.70 \\
             GPT-4o-May (IM 3) & 0.83 / 1.00 &  0.20 /    0.31 /       0.71 \\ \midrule \midrule
Llama3: SP Simplification & 0.66 / 0.88 &  0.25 /    0.44 /       0.73 \\
Llama3: Exclusion of Refs & 0.71 / 0.90 &  0.25 /    0.44 /       0.74 \\

Llama3: Comment First & 0.51 / 0.81 &  0.24 /    0.44 /       0.72 \\
\bottomrule
\end{tabular}
    
    \caption{Reliability tests results. ``QWK / AAR1'' and ``BLEU / ROUGE-L / BERTScore'' are used to measure score stability and comment similarity, respectively.}
    \label{tab:reliabilityResults}
\end{table}

\subsection{Reliability of LLM-generated Assessments} 

We evaluate the reliability of LLM-generated assessments across different realistic conditions that mirror potential real-world use cases. To prevent experimental confounding, we change only one condition at a time for a given LLM in a specific interaction mode, assuming that users tend to interact with their chosen LLM in a consistent manner.

First, we consider \textsc{gpt-4o-2024-08-06} (GPT-4o-Aug) in IM 1 with the default prompt setting from Section~\ref{sec:prompting} as the baseline. To test the effect of model variant, we run the same experiment but with \textsc{gpt-4o-2024-05-13} (GPT-4o-May). We also prompt GPT-4o-Aug while varying one of the four conditions in the default prompt setting (see Section~\ref{sec:prompting}) by (1) removing the helpful information from the system prompt, (2) excluding references in the input essays, (3) instructing LLMs to produce a comment before a score, or (4) setting temperature to 1 to increase output randomness. 

To ensure the comprehensiveness of our experiments, we prompt GPT-4o-May in IM 2 and IM 3 under default prompt setting to study the effect of model variant under other interaction modes. We also prompt Llama-3 in IM 1 changing the first three conditions in the default prompt setting mentioned in the last paragraph. The baselines here are GPT-4o-Aug and Llama-3 prompted under respective interaction modes from Section~\ref{sec:prompting}.

We use QWK and AAR1 and three widely adopted machine translation metrics, i.e., BLEU \citep{papineni2002bleu}, ROUGE-L \citep{lin-2004-rouge}, and BERTScore \citep{bert-score}, to evaluate the reliability of the generated scores and comments between contrastive condition pairs, respectively. 

The results in Table~\ref{tab:reliabilityResults} show that LLMs are capable of generating highly stable scores, with an AAR1 score at least 0.81 and mostly above 0.9 across all conditions. Their generated comments are also decently similar with BERTScore typically no lower than 0.67. A small-scale manual check and a correlation analysis performed in Appendix~\ref{app:furtherAnalyses} further verify the validity of BERTScore in measuring comment similarity.

\section{Conclusion\label{sec:conclusion}} 

This study provides evidence that LLMs can generate reasonably good and generally reliable multi-dimensional analytic assessments. Our findings highlight the promising role of LLMs in assessing academic English writing, especially for graduate-level literature reviews, which is a highly technical genre. In short, LLMs show strong pedagogical potential, benefiting both L2 learners and instructors for self-regulated learning or teaching assistance. We propose and validate a novel problem-focused evaluation framework, namely ProEval, to facilitate our analysis. Our stduy demonstrates that ProEval is time-
and cost-efficient, scalable, and reproducible, compared to manual judgments. It is also interpretable and fine-grained, compared to direct quality ratings.

Looking ahead, future studies could further characterize and compare the writing problems identified by human- versus LLM-generated comments, offering deeper qualitative insights. Additionally, it would be valuable to develop a metric grounded in our proposed framework that can directly compare the relative quality of two sets of comments. We release our corpus to support continued research in this area.

\section*{Acknowledgments}

Zhengxiang Wang, Veronika Makarova, and Zhi Li would like to thank Social Sciences and Humanities Research Council of Canada (SSHRC) for funding the writing project (``Collaborative development of written academic genre awareness by international graduate students") under the Insight Development Grants (430-2020-00179). They also appreciate three graduate students, i.e., Leslee G. Mann, Abdelrahman Alqudah, and Hanh Pham who expertly assessed the participants’ submitted writings, and the participants who participated in the project. 

Zhengxiang Wang and Owen Rambow were supported in part by funding from the Defense Advanced Research
Projects Agency (DARPA) under Contracts No.HR01121C0186, No. HR001120C0037, and PR No. HR0011154158. Any opinions, findings and conclusions or recommendations expressed in this
material are those of the authors and do not necessarily reflect the views of DARPA.

Zhengxiang Wang, Jordan Kodner, and Owen Rambow are grateful for the supports from the Institute for Advanced Computational Science (IACS) at Stony Brook University, in particular the free GPT access it provides. Zhengxiang Wang is supported by IACS's Junior Researcher Award since Fall 2024.

We thank Yongjun Zhang and the three anonymous reviewers for their valuable feedback. This work was presented at several venues, including \textit{All Things Language and Computation (ATLAC)} at Stony Brook University, the \textit{Mid-Atlantic Student Colloquium on AI, Language, and Learning (MASC-ALL)} at Penn State University, and the \textit{New England NLP Meeting Series (NENLP)} at Yale University. We are grateful for the insightful discussions and feedback received from the audiences at these events.

We thank Hannah Stortz for providing manual annotations for our study.

\section*{Limitations}

\paragraph{Generality of Findings} This study focuses on L2 graduate-level academic writing, specifically literature reviews in the humanities and social sciences. While this domain represents a significant subset of academic writing, the findings may not generalize to other genres (e.g., technical reports, creative writing) or proficiency levels (e.g., undergraduate or professional writers). Additionally, our study is limited to English, a high-resource language, which means our results may not be indicative of LLMs’ capabilities in other languages, particularly low-resource ones. Future research should explore the applicability of our findings across diverse writing contexts and linguistic backgrounds.

\paragraph{Weakness of Our Assumption About Feedback Quality} A key limitation of our approach is that it does not account for other factors that may influence the \textit{perceived} quality of a feedback comment, such as politeness (e.g., rude comments may not be well received) or the logical coherence of the argument (e.g., illogical comments could be misleading). However, this concern is less pronounced for LLM-generated feedback comments, as LLMs are trained to align with human preferences and social norms \citep{ouyang2022traininglanguagemodelsfollow}. Moreover, these factors could potentially be incorporated into our framework by adding additional steps focused on politeness and argumentation etc.

\paragraph{Indirect Evaluation of Feedback Quality} While our approach to measuring the general quality of LLM-generated assessments is intuitive and simple, it is inherently indirect. A large-scale manual evaluation remains necessary to more accurately assess and compare the quality of human- and LLM-generated multi-dimensional analytic assessments. Due to resource constraints, we leave this investigation to future studies.

\paragraph{Limited Validation and Reliability Testing} Due to time and resource constraints, we were unable to comprehensively validate our proposed feedback comment quality evaluation framework. As a result, we may have overlooked some potential issues with the framework or the LLM outputs. Similarly, the reliability assessments we conducted are limited, with only one factor being changed at a time in each evaluation. More extensive experiments are needed to further validate our claim that LLM-generated assessments are generally reliable and to explore the conditions influencing this reliability.

\section*{Ethical Considerations}

\paragraph{Corpus Creation} The research project that led to the construction of the corpus was ethically reviewed and received approval from the University of Saskatchewan for involving human participants. Participants provided informed consent to allow the use of their materials, with the option to withdraw at any time.

\paragraph{Human Annotations} We compensated the hired annotator at a rate of approximately US\$25 per hour, which exceeds the minimum wage in the region where the annotations took place.

\paragraph{Potential Biases in LLM Assessments} LLMs are trained on large-scale datasets that may contain inherent biases, which can be reflected in their assessments. For example, they might systematically favor certain writing styles, linguistic structures, or cultural conventions, leading to biased evaluations. However, we argue that in contexts where human assessments are not readily accessible, the benefits of LLM-generated feedback\,\---\,particularly for L2 learners\,\---\,may outweigh potential biases. Furthermore, bias mitigation strategies, such as improved prompting techniques or advancements in LLM development, could help reduce these concerns.

\bibliography{references}

\appendix

\section{Corpus\label{app:corpus}}

\subsection{Basic Corpus Statistics}

Table~\ref{tab:basicStatistics} provides the basic statistics of the corpus. Note that throughout this study, we use the default word tokenizer of NLTK to compute word counts. See: \url{https://www.nltk.org/api/nltk.tokenize.html}.

\begin{table}[]
    \centering
    \small
    \begin{tabular}{llllll}
    \toprule
    &  T1 & T2 & T3 & T4 & T5 \\ \midrule
    \# Essays     & 50 & 16 & 31 & 13 & 31 \\ 
    
    Avg WC (w/o refs) & 845 & 1169 & 926 & 1079 & 887 \\ 
    
    Avg WC (w/ refs)  & 1232 & 1583 & 1347 & 1666 & 1159 \\ \bottomrule
    \end{tabular}
    \caption{Basic statistics of the corpus. ``T'' in each column stands for ``Topic.'' ``WC'' means ``word count.''}
    \label{tab:basicStatistics}
\end{table}

\subsection{Details of the 5-Unit Tutorial Series}

Table~\ref{tab:details-of-the-5-unit-tutorial-series} presents details of the 5-unit tutorial series, including the themes, notions, activities, duration, and writing task for each unit.



\subsection{Assessment Criteria\label{app:criteria}}

The 9 assessment criteria/questions provided to human assessors are detailed in Table~\ref{tab:assessmentCriteria}.

\begin{table*}[]
    \centering
    \scriptsize
    
    \begin{tabular}{p{2cm}|p{3.5cm}|p{3cm}|p{1.5cm}|p{3.5cm}}
    \toprule
      \textbf{Unit}  & \textbf{Key notions} & \textbf{Activities} & \textbf{Duration} & \textbf{Writing task} \\ \midrule
      1. Genre of literature review   & Components in literature review writing, material selection, citation practices & Interactive e-book, Peer-review, Discussion forum, quiz & 3 weeks & Individual writing on the social consequences of legalized cannabis \\ \midrule

      2. Structure and logic in literature review  & Types of logic structure, terms and abbreviations, coherence, cohesion & Interactive e-book, Discussion forum, quiz & 2 weeks & Collaborative writing on Canadian linguistic landscape \\ \midrule

      3. Sentence structures  & Sentence structures and variety, nominalization, Phrase Bank and Swales’ CARS (Creating a Research Space) model & Interactive e-book, Peer-review, Discussion forum, quiz & 3 weeks & Individual writing on the pros and cons of online learning \\ \midrule

      4. Academic vocabulary  & Academic formulaic expressions and their functions & Interactive e-book, Discussion forum, quiz & 2 weeks & Collaborative writing on lessons from the COVID- 19 pandemic \\ \midrule

      5. Grammar of reported speech   & Direct vs. indirect speech, reporting verbs and expressions, verb tenses, modal verbs & Interactive e-book, Peer-review, Discussion forum, quiz & 3 weeks & Individual writing on pacifism, peace-making, or just/justifiable war \\ \bottomrule
      
    \end{tabular}
    \caption{Details of the 5-unit online tutorial series.}
    \label{tab:details-of-the-5-unit-tutorial-series}
\end{table*}

\begin{table*}[]
    \centering
    \footnotesize
    \begin{tabular}{p{2.75cm}p{3.75cm}p{7.75cm}}
      \toprule
      Aspect & Criterion & Question \\ \midrule
         
    \multirow{3}{4em}{Selection of materials and citation practices}   & 1. Material selection & On a scale of 10 (1: Very poor, 10: Excellent), how would you evaluate the author’s selection of source materials in terms of relevance, quality, and quantity of the materials? Note: ``If the draft has a noticeable issue regarding the number or the quality of the papers reviewed, please comment on the issue." \\ \cmidrule{2-3}
         
          & 2. Material integration \newline and citation & On a scale of 10 (1: Very poor, 10: Excellent), how would you evaluate the writing for its integration of source materials (e.g., clarity of presenting information) and citation practices (e.g., use of APA or other style in both in-text citations and reference list)? \\ \midrule

     \multirow{3}{4em}{Overall structure}     & 3. Quality of key components & On a scale of 10 (1: Very poor, 10: Excellent), how would you evaluate the writing for the quality or effectiveness of each component (i.e., Introduction, Body, and Conclusions)? Note: The introduction is expected to introduce a research area, identify issue(s), and/or state the significance of the issue(s). The body of literature review should present the relevant ideas or findings of the reviewed studies and/or identify a research gap. The conclusion(s) may identify research trends or controversies and highlight the contribution of this literature review. \\ \cmidrule{2-3}
         
          & 4. Logic of structure & On a scale of 10 (1: Very poor, 10: Excellent), how would you evaluate the logical structure of this literature review? \\ \cmidrule{2-3}

          & 5. Content and clarity of ideas & On a scale of 10 (1: Very poor, 10: Excellent), how would you evaluate the content and clarity of ideas expressed in this literature review? \\ \midrule
         
     \multirow{2}{4em}{Coherence and cohesion}     & 6. Coherence & On a scale of 10 (1: Very poor, 10: Excellent), how would you evaluate the literature review for the quality of coherence (e.g., the connectivity and the naturalness of the flow of ideas in this draft)? \\ \cmidrule{2-3}

          & 7. Cohesion & On a scale of 10 (1: Very poor, 10: Excellent), how would you evaluate the literature review for the use of connectors (e.g., ‘because,’ ‘therefore,’ ‘however,’ ‘likewise’, and ‘similarly’) to link sentences in this draft? \\ \midrule

       \multirow{2}{4em}{Grammar and vocabulary}    & 8. Grammatical and sentence structure & On a scale of 10 (1: Very poor, 10: Excellent), how would you evaluate the draft for grammatical accuracy, sentence length and sentence type variety? \\ \cmidrule{2-3}
         
          & 9. Academy vocabulary & On a scale of 10 (1: Very poor, 10: Excellent), how would you evaluate the draft for vocabulary quality (e.g., use of academic expressions, the correctness of word choice, the naturalness of collocations, the complexity of vocabulary, the use of stylistically acceptable vocabulary—not too colloquial, not excessively formal or not overusing terms)? \\ \bottomrule
         
    \end{tabular}
    \caption{The 9 assessment criteria/questions, reflecting 4 general aspects of writing quality.}
    \label{tab:assessmentCriteria}
\end{table*}

\section{Feedback Comment Quality Evaluation Framework \label{app:framework}}

\subsection{Implementation}

The framework is implemented using LLMs. More concretely, we used \textsc{gpt-4o-2024-11-20} for Problem Extraction and Problem Classification, and \textsc{gpt-4-turbo-2024-04-09} for Correction Relevance Check. An example implementation of our framework can be found in Table~\ref{tab:frameworkImpExample}.

Related prompts used for implementing our framework can be found in Appendix~\ref{app:promptFramework}.

\begin{table*}[]
    \centering
    \scriptsize
    
\begin{tabular}{p{3cm}|p{3cm}|p{4cm}|p{5cm}}
\toprule
             Comment &              Problem Extraction & Problem Classification & Correction Relevance Check \\
\midrule
The author has generally done a good job of integrating the source materials into the text, with clear summaries and explanations of the findings. However, \textcolor{cyan}{there are some areas where the citation practices could be improved. For example, some of the in-text citations are not formatted correctly (e.g., "Wilkinson ST, et al., 2015" should be "Wilkinson et al., 2015")}, and \textcolor{green!60!black}{there are some inconsistencies in the reference list (e.g., some sources have DOIs, while others do not)}. Additionally, \textcolor{purple}{the author could benefit from using more precise language when describing the findings of the studies, rather than relying on general statements.}  & \textcolor{cyan}{There are some areas where the citation practices could be improved. For example, some of the in-text citations are not formatted correctly (e.g., "Wilkinson ST, et al., 2015" should be "Wilkinson et al., 2015").} & 1. Yes. The excerpt refers to a specific part of the essay by mentioning the in-text citation "Wilkinson ST, et al., 2015" and how it should be formatted. \newline 2. Yes. The excerpt offers a suggestion by indicating that the in-text citation formatting should be corrected. \newline 3. Yes. The excerpt provides a concrete correction by specifying the correct format for the citation as "Wilkinson et al., 2015." \newline\newline Final answers: Yes, Yes, Yes  &                   1. Yes, the problem pointed out in the excerpt exists in the corresponding essay. The excerpt mentions that the citation "Wilkinson ST, et al., 2015" is incorrectly formatted and should be "Wilkinson et al., 2015". In the essay, the citation appears as "Wilkinson ST, et al., 2015", which is indeed not formatted correctly according to APA style guidelines that recommend omitting the first names and middle initials of authors in in-text citations. \newline\newline

2. Yes, the problem pointed out in the excerpt is relevant to the corresponding assessment question. The assessment question asks to evaluate the essay based on the integration of source materials and citation practices. The excerpt directly addresses an issue with citation practices, which is a key aspect of the assessment question. \newline\newline

3. Yes, the correction of the problem pointed out in the excerpt is correct. The suggested correction to change "Wilkinson ST, et al., 2015" to "Wilkinson et al., 2015" aligns with APA style guidelines, which recommend using only the surname of the first author followed by "et al." for works authored by multiple people, without including initials unless necessary to distinguish between authors with the same surname. \newline\newline

Final answers: Yes, Yes, Yes \\ \noalign{\vskip 6pt} \cline{2-4} \noalign{\vskip 6pt}
 & \textcolor{green!60!black}{There are some inconsistencies in the reference list (e.g., some sources have DOIs, while others do not).} & 1. No - The excerpt does not refer to a specific part of the essay, as it does not point to a particular location or provide a specific example from the reference list. \newline 2. Yes - The excerpt suggests that the student should address the inconsistencies in the reference list, such as ensuring uniformity in the inclusion of DOIs. \newline 3. No - The excerpt does not provide a concrete correction, as it does not specify exactly how to fix the inconsistencies or what the corrected references should look like. \newline \newline Final answers: No, Yes, No &    \\ \noalign{\vskip 6pt} \cline{2-4} \noalign{\vskip 6pt}

 & \textcolor{purple}{The author could benefit from using more precise language when describing the findings of the studies, rather than relying on general statements.} & 1. No - The excerpt does not refer to a specific part of the essay. It does not mention a concrete location or provide any referencable details about where the issue occurs. \newline 2. Yes - The excerpt offers a general suggestion for improvement by advising the author to use more precise language when describing the findings of the studies. \newline 3. No - The excerpt does not provide a concrete correction or specific example of how to fix the issue. \newline\newline Final answers: No, Yes, No &                \\
\bottomrule
\end{tabular}
    
    \caption{An example implementation of our framework ProEval on a feedback comment on the second assessment criteria, i.e., material integration \& citation, which is about citation practices. ProEval correctly identifies three problems (color-coded) identified in the comment and performs Problem Classification on each one of them, of which only the first problem offers a concrete correction. The Correction Relevance Check is thus only performed on the first extracted problem.}
    \label{tab:frameworkImpExample}
\end{table*}

\subsection{Annotation}

\paragraph{Guidelines} Table~\ref{tab:problemCharacterizations} provides explanations and examples of what is considered as a problem for Problem Extraction, and the three characteristics relevant to Problem Classification: whether an extracted problem (1) refers to a specific part of the essay, (2) provides a suggestion (general or specific), and (3) offers a concrete correction.

\paragraph{Samples for Problem Extraction} We employed stratified sampling to randomly select 100 human-generated feedback comments and 108 LLM-generated feedback comments. In total, there are 208 comments for manual annotations.

For LLM-generated comments, half of them were generated under Interaction Mode 1 and the other half under Interaction Modes 2 and 3. Comments from Interaction Modes 2 and 3 were sampled together to reduce manual annotation effort, as these comments tend to be lengthy. The sampling covered the 9 assessment criteria, with 2 comments from each of the 3 LLMs used, resulting in 9 * 3 * 2 = 54 comments from Interaction Mode 1 and another 54 comments from the combined Interaction Modes 2 and 3.

\paragraph{Samples for Problem Classification} We randomly sampled 100 problems extracted from both human- and LLM-generated comments, resulting in 200 problems for annotations. 

Since the distribution of extracted problems across the nine assessment criteria are highly skewed, we ensured that there were at least 5 problems for each assessment criterion.

\paragraph{Problem Extraction} For each feedback comment, the two annotators were provided with LLM-extracted problems and asked to identify the number of correctly extracted problems (true positives), the number of incorrectly extracted problems (false positives), and the number of problems not extracted (false negatives). The number of true negatives is always set to 0, as there is no negative prediction in problem extraction.

A problem is considered correctly extracted if the LLM output contains the exact or paraphrased problem stated or implied in the feedback comment. It is acceptable if additional information relevant to the problem, such as elaborations, suggestions, clarifying questions, or quoted text from the assessed essay, is not included in the LLM-identified problems, which appears to be uncommon based on our annotations. However, if the problem and relevant additional information are extracted as separate problems, only the stated or implied problem is counted as a true positive, and the relevant information is treated as a false positive. This over-segmentation is the primary source of errors in LLM-extracted problems.

\paragraph{Problem Classification} For each extracted problem, the two annotators were asked to answer the three classification problems based on Table~\ref{tab:problemCharacterizations}.

\begin{table*}[]
    \centering
    \footnotesize
    \begin{tabular}{p{2.75cm}p{4cm}p{7.5cm}}
      \toprule
      \textbf{Characteristic} & \textbf{Explanation} & \textbf{Examples} \\ \midrule

      If a problem is stated or implied in a comment & A problem is any writing-related issue that affects the quality of the writing, such as citation errors, logical flaws, coherence issues, grammatical mistakes, or inappropriate word choices, among others. The problem can be mentioned or implied in a given comment. & \underline{Positive Examples}
      \begin{itemize}
          \itemsep-0.25em
          \item Specify what the abbreviation ``THC'' stands for. (Implied problem: ``THC'' is unspecified)
          \item There was a redundant use of ``the legalization of cannabis''.	
      \end{itemize} \underline{Negative Examples} \begin{itemize}
          \itemsep-0.25em
          \item Great grammatical skills, well done!
          \item Final references are well formatted. In-text references are well integrated.	
      \end{itemize} \\ \midrule

      If a problem points to a specific part of the essay & A specific part refers to a part of the essay that is easily locatable. (1) It can be a specific word, phrase, sentence, paragraph, reference etc. used in the essay. (2) It can be a concrete location, such as ``sentence 2 in paragraph 2,'' ``in paragraph 6,'' ``the first citation,'' or ``the first sentence of the paper'' and so on. (3) A less concrete location, such as ``the introduction,'' or ``the conclusion,'' is also considered a specific part if it is accompanied by some referenceable details. & \underline{Positive Examples}
      \begin{itemize}
          \itemsep-0.25em
          \item In Paragraph 2, the word ``decay'' is likely a mistake and should be replaced with ``decade''.	
          \item The sentence ``This theory still is under debate even with many authors provide a justification for that'' contains a grammatical error. The verb ``provide'' should be corrected to ``providing.''	
      \end{itemize} \underline{Negative Examples} \begin{itemize}
          \itemsep-0.25em
          \item Some of the sentences are a bit too long and fall apart a little.	
          \item Your paper would benefit from the use of expressions such as ``as a result'' or ``the result'' where cause and consequence are important.	
      \end{itemize} \\ \midrule

      If a problem offers some form of suggestions, general or specific & A suggestion indicates or implies ares of improvement. If the problem only contains a problem statement and it is unclear what direction the student should take to improve it, then there is no suggestion. A concrete correction is always considered a suggestion. & \underline{Positive Examples}
      \begin{itemize}
          \itemsep-0.25em
          \item Some sentences could be a bit shorter.	
          \item The use of a topic sentence for each paragraph in the main body could be improved.		
      \end{itemize} \underline{Negative Examples} \begin{itemize}
          \itemsep-0.25em
          \item The beginning of the literature review could be changed slightly.
          \item The first sentence of the paper is confusing.	
      \end{itemize} \\ \midrule

      If a problem provides a concrete correction for an identified writing issue & A concrete correction is something that can be directly applied to an essay to fix a writing problem. Corrections should not require thinking to implement, i.e. text that can be copy-pasted, or actions that can be taken following an instruction (e.g., capitalize the first letter). & \underline{Positive Examples}
      \begin{itemize}
          \itemsep-0.25em
          \item The citation ``(Toronto Star December 2016)'' should be revised to ``(Toronto Star 2016)'' to align with proper citation practices.	
          \item ``The advance of technologies'' should be corrected to ``the advancement of technologies''.		
      \end{itemize} \underline{Negative Examples} \begin{itemize}
          \itemsep-0.25em
          \item The significance of South Australian policy is unclear, as it is the first citation and the only one in the Introduction.	
          \item The conclusion is a little too short.	
      \end{itemize}  \\
      
      \bottomrule   
    \end{tabular}
    \caption{Explanations and illustrative examples of ``problems'' and their characterizations.}
    \label{tab:problemCharacterizations}
\end{table*}

\subsection{Correction Relevance Check\label{app:relevanceCheck}}

Table~\ref{tab:overallRelevanceCheckResults} demonstrates that comments generated by both humans and LLMs are overall highly relevant. However, human-generated comments tend to exhibit slightly lower relevance—either broadly or strictly—compared to those generated by LLMs. 

We conducted a small-scale error analysis to investigate the reasons behind the 8\%, 15\%, and 9\% of human-identified problems that GPT-4 incorrectly classified as not present in the essays, not adhering to the assessment criteria, and being incorrect, respectively. 

\paragraph{Problems not Present in Essays} We randomly selected 10 problems identified by GPT-4 as not present in the assessed essays. Upon reviewing each human-identified problem in the original essay, we found that 6 of these problems were indeed present, while 4 were not. Of the 4 problems that did not exist in the essays, 3 appeared to be misassigned comments (2 of these 3 were extracted from the same comment), while the remaining one seemed to be an assessor error. Among the 6 problems that GPT-4 misclassified, 4 were due to GPT-4 misunderstanding the identified problems, 1 was due to GPT-4 failing to locate a quoted word in the essay, and 1 was because GPT-4 mistakenly deemed the identified problem not to be a problem, despite its presence in the essay.

\paragraph{Problems not Adherent to the Assessment Criteria} We randomly selected 10 problems identified by GPT-4 as not adhering to the assessment criteria. Of these, 9 were related to C8 (grammar \& sentence structure), and 1 was related to C9 (academic vocabulary). Our manual validation showed that 7 of the problems were less related to grammar and sentence structure but more related to word choice or clarity of expression. The remaining 3 were misclassified by GPT-4, mostly due to its requirement that problems be explicitly related to both grammar and sentence structure in order to adhere to C8.

\paragraph{Correction being Incorrect} We randomly selected 10 problems containing corrections identified by GPT-4 as incorrect. We found that 5 of these problems involved accurate corrections, all related to grammar. There were 2 corrections proposed to be suggestions and 3 corrections that require subjective judgments to determine their correctness. 

\paragraph{Remarks} Based on this error analysis, we can attributed the discrepancy in relevance to two primary reasons: (1) human comments often include (inconsistent use of) diacritics that complicate problem extraction and characterization, and (2) human assessors may occasionally deviate from instructions, providing corrections unrelated to the assessment question. These issues are less frequent in LLM-generated comments, which benefit from their strong adherence to instructions and the ability to handle extended context windows. That said, both human- and LLM-identified problems are highly relevant.

\begin{table*}[]
    \centering
    \small

\begin{tabular}{lrrrrr}
\toprule
         Assessor &  In Essay &  In Question &  Is Correct &  Broadly Relevant &  Strictly Revelant \\
\midrule
               Human B &     87.9 &        79.4 &       85.1 &           84.4 &            72.8 \\
               Human C &     94.9 &        91.8 &       94.5 &           93.8 &            89.0 \\
               Human F &     96.3 &        86.7 &       91.4 &           90.9 &            82.3 \\ \midrule
    GPT-4o (IM 1) &    100.0 &       100.0 &      100.0 &          100.0 &           100.0 \\
Gemini-1.5 (IM 1) &     95.6 &        99.6 &       98.0 &           95.6 &            95.2 \\
   Llama-3 (IM 1) &     97.8 &        97.8 &       97.8 &           97.8 &            97.8 \\ \midrule
    GPT-4o (IM 2) &     99.6 &       100.0 &      100.0 &           99.6 &            99.6 \\
Gemini-1.5 (IM 2) &     98.3 &        98.8 &       97.5 &           97.1 &            96.6 \\
   Llama-3 (IM 2) &     94.7 &        96.2 &       96.2 &           94.4 &            92.5 \\ \midrule
    GPT-4o (IM 3) &    100.0 &        99.5 &       99.8 &           99.8 &            99.2 \\
Gemini-1.5 (IM 3) &     98.8 &        97.8 &       99.0 &           98.8 &            96.8 \\
   Llama-3 (IM 3) &     98.7 &        98.7 &       98.5 &           98.5 &            97.5 \\
\bottomrule
\end{tabular}
    
    \caption{Overall Correction Relevance Check results (\%), representing the percentage of instances each attribute is true for corrections made by an assessor. ``In Essay'': whether the problem indicated in the correction exists in the essay. ``In Question'': whether the correction relates to the assessment question. ``Is Correct'': whether the correction is correct. ``Broadly Relevant'': applicable when both ``In Essay'' and ``Is Correct'' are true. ``Strictly Revelant'': applicable when both ``Broadly Relevant'' and ``In Question'' are true.}
    \label{tab:overallRelevanceCheckResults}
\end{table*}

\section{Results\label{app:results}}

\subsection{Number of Co-Assessed Essays\label{app:numberCoAssessed}}

Table~\ref{tab:number_CoAssessedEssays} shows the number of essays co-assessed by different assessor pairs.

\begin{table*}[]
    \centering
    \small

\begin{tabular}{lrrrrrr}
\toprule
{} &  Human B &  Human C &  Human F &  Llama-3 (IM 1) &  Gemini-1.5 (IM 1) &  Llama-3 (IM 2) \\
\midrule
Human B           &      141 &       93 &      106 &             140 &                140 &             139 \\
Human C           &       93 &       93 &       78 &              93 &                 92 &              93 \\
Human F           &      106 &       78 &      106 &             106 &                105 &             106 \\
Llama-3 (IM 1)    &      140 &       93 &      106 &             140 &                139 &             139 \\
Gemini-1.5 (IM 1) &      140 &       92 &      105 &             139 &                140 &             138 \\
Llama-3 (IM 2)    &      139 &       93 &      106 &             139 &                138 &             139 \\
\bottomrule
\end{tabular}
    
    \caption{Number of essays co-assessed by different assessor pairs. We only show three LLMs, which failed to generate assessments for all 141 essays in the corpus for some reason (e.g., content moderation, exceeding context window). We omit the other LLMs since they assessed all essays and the numbers of essays they co-assessed with the five assessors in the table excluding human B are identical to those between human B and those five assessors. The number of essays the omitted LLMs and human B co-assessed is always 141.}
    \label{tab:number_CoAssessedEssays}
\end{table*}

\subsection{Scores\label{app:scores}}

\paragraph{Scoring Ranges} Table~\ref{tab:descStats} summarizes the scoring ranges, in the form of means and standard deviations for each assessment criterion, as produced by three human assessors and the three LLMs under three interaction modes.

\paragraph{Full QWK/AAR1} Table~\ref{tab:QWK_full} presents the full results for Quadratic Weighted Kappa (QWK) and Table~\ref{tab:AAR1_full} presents the full results for AAR1. 

\paragraph{Inconsistencies in Scoring by Human Assessors} First, there is an instance in the corpus, where assessor B accidentally assessed the same essay twice on separate days.\footnote{Four days apart and assessor B had no access to their earlier assessments.} While assessor B provided identical scores for 5 out of the 9 assessment criteria, discrepancies of 1 point occurred for the remaining 4 criteria, with scores alternating between (8, 7), (8, 7), (4, 5), and (7, 8). 

Second, we observe that human assessors assigned different scores to identical or similar comments, mostly within 1-point differences. For example, assessor F gave the same comment ``Decent number of citations'' three times but assigned three different scores: 6, 7, and 8. Similarly, assessor C assigned scores of 7 and 8 to the comment ``Appropriate use of connectors.'' However, when the same comment is repeated, scores tend to be very close, typically within one point. For instance, assessor A assigned a score of 8 to the comment ``Great use of academic words and formal tone'' five times, and there was only one more instance where the score was 9.

\begin{table*}[]
    \small
    \centering

\begin{tabular}{llllllllll}
\toprule
         Assessor &             C1 &             C2 &             C3 &             C4 &             C5 &             C6 &             C7 &             C8 &             C9 \\
\midrule
             Human B & 6.7{\tiny±0.9} & 6.5{\tiny±1.2} & 7.5{\tiny±1.2} & 7.7{\tiny±1.1} & 7.7{\tiny±1.1} & 7.6{\tiny±1.1} & 7.3{\tiny±1.1} & 7.2{\tiny±1.1} & 7.5{\tiny±1.1} \\
               
              Human C & 7.8{\tiny±1.3} & 7.6{\tiny±1.3} & 7.9{\tiny±1.0} & 7.8{\tiny±1.3} & 7.8{\tiny±1.1} & 7.9{\tiny±1.1} & 8.1{\tiny±0.9} & 7.7{\tiny±1.1} & 8.2{\tiny±0.9} \\ 

              Human F & 7.0{\tiny±1.0} & 6.6{\tiny±1.0} & 6.9{\tiny±0.9} & 7.0{\tiny±0.8} & 7.1{\tiny±0.8} & 7.1{\tiny±0.8} & 7.2{\tiny±0.8} & 7.3{\tiny±0.7} & 7.0{\tiny±0.8} \\ \midrule
               
    GPT-4o (IM 1) & 7.4{\tiny±0.7} & 6.4{\tiny±0.7} & 5.7{\tiny±0.8} & 5.7{\tiny±0.9} & 6.3{\tiny±0.7} & 5.4{\tiny±0.7} & 5.5{\tiny±0.8} & 6.4{\tiny±0.9} & 6.7{\tiny±0.8} \\
    GPT-4o (IM 2) & 6.9{\tiny±0.7} & 6.0{\tiny±0.8} & 6.0{\tiny±0.8} & 5.6{\tiny±1.1} & 6.2{\tiny±0.8} & 5.4{\tiny±0.9} & 4.9{\tiny±0.8} & 6.2{\tiny±0.9} & 6.8{\tiny±0.9} \\
    GPT-4o (IM 3) & 6.9{\tiny±0.7} & 6.4{\tiny±0.7} & 6.0{\tiny±0.7} & 6.2{\tiny±0.7} & 6.4{\tiny±0.6} & 6.1{\tiny±0.7} & 6.0{\tiny±0.7} & 6.7{\tiny±0.7} & 6.8{\tiny±0.6} \\ \midrule
    
Gemini-1.5 (IM 1) & 6.3{\tiny±0.8} & 5.4{\tiny±0.7} & 5.5{\tiny±0.7} & 5.5{\tiny±1.0} & 6.0{\tiny±0.8} & 4.9{\tiny±0.8} & 4.5{\tiny±0.9} & 5.7{\tiny±0.8} & 6.1{\tiny±0.8} \\
Gemini-1.5 (IM 2) & 6.4{\tiny±0.6} & 6.3{\tiny±0.9} & 5.5{\tiny±0.7} & 5.8{\tiny±0.8} & 6.0{\tiny±0.5} & 5.4{\tiny±0.7} & 5.2{\tiny±0.8} & 6.4{\tiny±0.6} & 6.5{\tiny±0.6} \\
Gemini-1.5 (IM 3) & 6.4{\tiny±0.6} & 5.8{\tiny±0.6} & 5.5{\tiny±0.6} & 5.6{\tiny±0.5} & 5.7{\tiny±0.5} & 5.5{\tiny±0.6} & 5.4{\tiny±0.5} & 6.0{\tiny±0.6} & 6.1{\tiny±0.6} \\ \midrule

   Llama-3 (IM 1) & 7.5{\tiny±0.5} & 7.4{\tiny±0.7} & 6.4{\tiny±0.9} & 6.4{\tiny±1.2} & 7.1{\tiny±0.7} & 6.2{\tiny±0.8} & 5.2{\tiny±0.7} & 7.8{\tiny±0.5} & 7.1{\tiny±0.7} \\
   Llama-3 (IM 2) & 7.2{\tiny±0.6} & 6.8{\tiny±1.0} & 6.1{\tiny±1.1} & 6.4{\tiny±1.4} & 6.7{\tiny±1.1} & 6.2{\tiny±1.4} & 4.9{\tiny±1.4} & 7.3{\tiny±0.9} & 7.2{\tiny±0.8} \\
   Llama-3 (IM 3) & 7.2{\tiny±0.5} & 6.9{\tiny±0.5} & 6.4{\tiny±0.6} & 6.7{\tiny±0.6} & 6.8{\tiny±0.4} & 6.7{\tiny±0.6} & 5.9{\tiny±0.6} & 6.8{\tiny±0.4} & 6.8{\tiny±0.5} \\
\bottomrule
\end{tabular}

    \caption{Means and standard deviations of scores assigned by three human assessors and three LLMs prompted under three interaction modes (IM), denoted by ``IM'' in parentheses. C1: Material selection. C2: Material integration and citation; C3: Quality of key components. C4: Logic of structure. C5: Content and clarity of ideas. C6: Coherence (flow of ideas). C7: Cohesion (use of connectors). C8: Grammar and sentence structure. C9: Academic vocabulary.}
    \label{tab:descStats}
\end{table*}

\begin{table*}[]
    \footnotesize
    \centering

\begin{tabular}{lrrrrrrrrrr}
\toprule
                            Assessor &   C1 &   C2 &   C3 &    C4 &   C5 &   C6 &   C7 &   C8 &   C9 &  Overall \\
\midrule \midrule
                             Human B vs. Human F & 0.36 & 0.32 & 0.18 &  0.12 & 0.11 & 0.09 & 0.20 & 0.24 & 0.26 &     0.25 \\
                             Human B vs. Human C & 0.41 & 0.39 & 0.34 &  0.36 & 0.43 & 0.51 & 0.40 & 0.36 & 0.43 &     0.41 \\
                             Human F vs. Human C & 0.52 & 0.29 & 0.29 &  0.24 & 0.23 & 0.24 & 0.17 & 0.28 & 0.20 &     0.30 \\ \midrule \midrule
                             
                 Human B vs. GPT-4o (IM 1) & 0.23 & 0.29 & 0.08 &  0.06 & 0.11 & 0.05 & 0.06 & 0.25 & 0.10 &     0.03 \\
                 Human B vs. GPT-4o (IM 2) & 0.33 & 0.20 & 0.08 &  0.05 & 0.12 & 0.05 & 0.06 & 0.20 & 0.15 &     0.06 \\
                 Human B vs. GPT-4o (IM 3) & 0.26 & 0.30 & 0.09 &  0.07 & 0.14 & 0.07 & 0.09 & 0.28 & 0.18 &     0.10 \\ [0.15cm]
                 
             Human B vs. Gemini-1.5 (IM 1) & 0.29 & 0.15 & 0.08 &  0.07 & 0.10 & 0.04 & 0.05 & 0.12 & 0.08 &     0.07 \\
             Human B vs. Gemini-1.5 (IM 2) & 0.25 & 0.22 & 0.08 &  0.05 & 0.05 & 0.05 & 0.05 & 0.16 & 0.10 &     0.04 \\
             Human B vs. Gemini-1.5 (IM 3) & 0.25 & 0.18 & 0.06 &  0.06 & 0.06 & 0.04 & 0.05 & 0.09 & 0.03 &     0.04 \\ [0.15cm]
             
                Human B vs. Llama-3 (IM 1) & 0.10 & 0.04 & 0.07 & -0.03 & 0.03 & 0.01 & 0.03 & 0.06 & 0.08 &    -0.06 \\
                Human B vs. Llama-3 (IM 2) & 0.27 & 0.23 & 0.16 &  0.14 & 0.22 & 0.11 & 0.06 & 0.25 & 0.13 &     0.10 \\
                Human B vs. Llama-3 (IM 3) & 0.26 & 0.18 & 0.07 &  0.13 & 0.14 & 0.14 & 0.07 & 0.09 & 0.06 &     0.07 \\ \midrule
                
                 Human C vs. GPT-4o (IM 1) & 0.36 & 0.28 & 0.10 &  0.12 & 0.17 & 0.07 & 0.03 & 0.22 & 0.15 &     0.13 \\
                 Human C vs. GPT-4o (IM 2) & 0.27 & 0.21 & 0.14 &  0.07 & 0.15 & 0.05 & 0.04 & 0.20 & 0.16 &     0.11 \\
                 Human C vs. GPT-4o (IM 3) & 0.23 & 0.25 & 0.09 &  0.13 & 0.19 & 0.08 & 0.06 & 0.30 & 0.17 &     0.15 \\ [0.15cm]
                 
             Human C vs. Gemini-1.5 (IM 1) & 0.19 & 0.11 & 0.09 &  0.11 & 0.14 & 0.06 & 0.03 & 0.10 & 0.09 &     0.08 \\
             Human C vs. Gemini-1.5 (IM 2) & 0.12 & 0.21 & 0.08 &  0.05 & 0.08 & 0.06 & 0.04 & 0.15 & 0.10 &     0.08 \\
             Human C vs. Gemini-1.5 (IM 3) & 0.12 & 0.15 & 0.08 &  0.07 & 0.08 & 0.07 & 0.01 & 0.11 & 0.05 &     0.07 \\ [0.15cm]
             
                Human C vs. Llama-3 (IM 1) & 0.24 & 0.16 & 0.09 &  0.08 & 0.21 & 0.08 & 0.02 & 0.22 & 0.10 &     0.06 \\
                Human C vs. Llama-3 (IM 2) & 0.27 & 0.36 & 0.11 &  0.19 & 0.28 & 0.10 & 0.04 & 0.43 & 0.13 &     0.14 \\
                Human C vs. Llama-3 (IM 3) & 0.27 & 0.30 & 0.10 &  0.15 & 0.17 & 0.16 & 0.06 & 0.20 & 0.12 &     0.14 \\ \midrule
                
                 Human F vs. GPT-4o (IM 1) & 0.44 & 0.32 & 0.24 &  0.17 & 0.22 & 0.09 & 0.07 & 0.14 & 0.26 &     0.17 \\
                 Human F vs. GPT-4o (IM 2) & 0.51 & 0.30 & 0.36 &  0.17 & 0.25 & 0.12 & 0.06 & 0.10 & 0.27 &     0.17 \\
                 Human F vs. GPT-4o (IM 3) & 0.47 & 0.29 & 0.25 &  0.29 & 0.25 & 0.19 & 0.07 & 0.14 & 0.32 &     0.24 \\ [0.15cm] 
                 
             Human F vs. Gemini-1.5 (IM 1) & 0.37 & 0.16 & 0.22 &  0.11 & 0.18 & 0.08 & 0.03 & 0.05 & 0.13 &     0.11 \\
             Human F vs. Gemini-1.5 (IM 2) & 0.29 & 0.16 & 0.14 &  0.14 & 0.10 & 0.09 & 0.05 & 0.12 & 0.21 &     0.11 \\
             Human F vs. Gemini-1.5 (IM 3) & 0.29 & 0.12 & 0.17 &  0.14 & 0.09 & 0.09 & 0.05 & 0.03 & 0.09 &     0.10 \\ [0.15cm]
             
                Human F vs. Llama-3 (IM 1) & 0.32 & 0.07 & 0.28 &  0.24 & 0.18 & 0.19 & 0.04 & 0.10 & 0.23 &     0.13 \\
                Human F vs. Llama-3 (IM 2) & 0.50 & 0.18 & 0.23 &  0.22 & 0.21 & 0.22 & 0.05 & 0.07 & 0.26 &     0.16 \\
                Human F vs. Llama-3 (IM 3) & 0.50 & 0.21 & 0.27 &  0.35 & 0.19 & 0.25 & 0.04 & 0.05 & 0.12 &     0.18 \\ \midrule \midrule
                
    GPT-4o (IM 1) vs. Llama-3 (IM 1) & 0.59 & 0.01 & 0.35 &  0.30 & 0.28 & 0.20 & 0.36 & 0.07 & 0.41 &     0.45 \\
 GPT-4o (IM 1) vs. Gemini-1.5 (IM 1) & 0.33 & 0.38 & 0.65 &  0.64 & 0.59 & 0.58 & 0.34 & 0.51 & 0.51 &     0.60 \\
Llama-3 (IM 1) vs. Gemini-1.5 (IM 1) & 0.23 & 0.01 & 0.27 &  0.23 & 0.19 & 0.14 & 0.28 & 0.04 & 0.26 &     0.30 \\ [0.15cm]

 GPT-4o (IM 2) vs. Gemini-1.5 (IM 2) & 0.49 & 0.39 & 0.48 &  0.59 & 0.53 & 0.56 & 0.48 & 0.56 & 0.47 &     0.64 \\
    GPT-4o (IM 2) vs. Llama-3 (IM 2) & 0.62 & 0.33 & 0.57 &  0.47 & 0.57 & 0.46 & 0.52 & 0.37 & 0.52 &     0.60 \\
Gemini-1.5 (IM 2) vs. Llama-3 (IM 2) & 0.33 & 0.32 & 0.36 &  0.35 & 0.27 & 0.36 & 0.30 & 0.30 & 0.23 &     0.47 \\ [0.15cm]

    Llama-3 (IM 3) vs. GPT-4o (IM 3) & 0.56 & 0.40 & 0.48 &  0.53 & 0.36 & 0.46 & 0.55 & 0.56 & 0.58 &     0.58 \\
Llama-3 (IM 3) vs. Gemini-1.5 (IM 3) & 0.33 & 0.21 & 0.28 &  0.28 & 0.15 & 0.24 & 0.30 & 0.24 & 0.24 &     0.33 \\
 GPT-4o (IM 3) vs. Gemini-1.5 (IM 3) & 0.49 & 0.50 & 0.50 &  0.44 & 0.34 & 0.49 & 0.38 & 0.41 & 0.35 &     0.52 \\
\bottomrule
\end{tabular}
    
    \caption{Full QWK (Quadratic Weighted Kappa) results between all assessor pairs, evaluated at the level of each assessment criterion and the whole essay (``Overall''). C1: Material selection. C2: Material integration and citation; C3: Quality of key components. C4: Logic of structure. C5: Content and clarity of ideas. C6: Coherence (flow of ideas). C7: Cohesion (use of connectors). C8: Grammar and sentence structure. C9: Academic vocabulary.}
    \label{tab:QWK_full}
\end{table*}
\begin{table*}[]
    \footnotesize
    \centering

\begin{tabular}{lrrrrrrrrrr}
\toprule
                            Assessor &   C1 &   C2 &   C3 &   C4 &   C5 &   C6 &   C7 &   C8 &   C9 &  Overall \\
\midrule \midrule
                             Human B vs. Human F & 0.90 & 0.77 & 0.73 & 0.69 & 0.75 & 0.75 & 0.85 & 0.86 & 0.79 &     0.79 \\
                             Human B vs. Human C & 0.58 & 0.58 & 0.70 & 0.80 & 0.82 & 0.86 & 0.75 & 0.76 & 0.80 &     0.74 \\
                             Human F vs. Human C & 0.73 & 0.54 & 0.60 & 0.64 & 0.68 & 0.69 & 0.65 & 0.74 & 0.59 &     0.65 \\ \midrule \midrule
                             
                 Human B vs. GPT-4o (IM 1) & 0.89 & 0.84 & 0.33 & 0.27 & 0.46 & 0.25 & 0.36 & 0.74 & 0.60 &     0.53 \\
                 Human B vs. GPT-4o (IM 2) & 0.94 & 0.76 & 0.42 & 0.30 & 0.43 & 0.27 & 0.17 & 0.70 & 0.68 &     0.52 \\
                 Human B vs. GPT-4o (IM 3) & 0.94 & 0.85 & 0.37 & 0.41 & 0.50 & 0.40 & 0.54 & 0.84 & 0.74 &     0.62 \\ [0.15cm]
                 
             Human B vs. Gemini-1.5 (IM 1) & 0.88 & 0.59 & 0.27 & 0.24 & 0.33 & 0.13 & 0.12 & 0.46 & 0.42 &     0.38 \\
             Human B vs. Gemini-1.5 (IM 2) & 0.94 & 0.75 & 0.28 & 0.32 & 0.30 & 0.23 & 0.29 & 0.71 & 0.62 &     0.49 \\
             Human B vs. Gemini-1.5 (IM 3) & 0.94 & 0.72 & 0.29 & 0.25 & 0.23 & 0.24 & 0.32 & 0.60 & 0.45 &     0.45 \\ [0.15cm]
             
                Human B vs. Llama-3 (IM 1) & 0.84 & 0.76 & 0.49 & 0.54 & 0.74 & 0.46 & 0.27 & 0.79 & 0.74 &     0.63 \\
                Human B vs. Llama-3 (IM 2) & 0.88 & 0.73 & 0.38 & 0.50 & 0.65 & 0.45 & 0.25 & 0.77 & 0.71 &     0.59 \\ 
                Human B vs. Llama-3 (IM 3) & 0.89 & 0.84 & 0.50 & 0.57 & 0.70 & 0.63 & 0.46 & 0.87 & 0.72 &     0.69 \\ \midrule
                
                 Human C vs. GPT-4o (IM 1) & 0.74 & 0.54 & 0.20 & 0.22 & 0.46 & 0.10 & 0.13 & 0.45 & 0.42 &     0.36 \\
                 Human C vs. GPT-4o (IM 2) & 0.65 & 0.44 & 0.32 & 0.17 & 0.38 & 0.15 & 0.03 & 0.43 & 0.54 &     0.35 \\
                 Human C vs. GPT-4o (IM 3) & 0.62 & 0.57 & 0.27 & 0.39 & 0.48 & 0.33 & 0.24 & 0.66 & 0.57 &     0.46 \\ [0.15cm]
                 
             Human C vs. Gemini-1.5 (IM 1) & 0.42 & 0.21 & 0.17 & 0.15 & 0.30 & 0.05 & 0.03 & 0.29 & 0.23 &     0.21 \\
             Human C vs. Gemini-1.5 (IM 2) & 0.47 & 0.51 & 0.16 & 0.26 & 0.35 & 0.17 & 0.08 & 0.53 & 0.45 &     0.33 \\
             Human C vs. Gemini-1.5 (IM 3) & 0.47 & 0.32 & 0.14 & 0.18 & 0.22 & 0.15 & 0.09 & 0.38 & 0.24 &     0.24 \\ [0.15cm]
             
                Human C vs. Llama-3 (IM 1) & 0.82 & 0.73 & 0.49 & 0.47 & 0.74 & 0.46 & 0.03 & 0.86 & 0.68 &     0.59 \\
                Human C vs. Llama-3 (IM 2) & 0.71 & 0.67 & 0.24 & 0.42 & 0.57 & 0.31 & 0.13 & 0.84 & 0.72 &     0.51 \\
                Human C vs. Llama-3 (IM 3) & 0.71 & 0.75 & 0.44 & 0.57 & 0.68 & 0.57 & 0.18 & 0.68 & 0.58 &     0.57 \\ \midrule
                
                 Human F vs. GPT-4o (IM 1) & 0.94 & 0.82 & 0.61 & 0.56 & 0.80 & 0.37 & 0.41 & 0.71 & 0.87 &     0.68 \\
                 Human F vs. GPT-4o (IM 2) & 0.96 & 0.80 & 0.79 & 0.42 & 0.71 & 0.32 & 0.17 & 0.65 & 0.86 &     0.63 \\
                 Human F vs. GPT-4o (IM 3) & 0.94 & 0.85 & 0.71 & 0.83 & 0.87 & 0.70 & 0.60 & 0.86 & 0.93 &     0.81 \\ [0.15cm] 
                 
             Human F vs. Gemini-1.5 (IM 1) & 0.77 & 0.63 & 0.56 & 0.50 & 0.64 & 0.19 & 0.10 & 0.45 & 0.72 &     0.51 \\
             Human F vs. Gemini-1.5 (IM 2) & 0.80 & 0.76 & 0.54 & 0.65 & 0.62 & 0.43 & 0.27 & 0.76 & 0.89 &     0.64 \\
             Human F vs. Gemini-1.5 (IM 3) & 0.80 & 0.75 & 0.53 & 0.57 & 0.53 & 0.44 & 0.35 & 0.57 & 0.75 &     0.59 \\ [0.15cm]
             
                Human F vs. Llama-3 (IM 1) & 0.87 & 0.68 & 0.79 & 0.73 & 0.89 & 0.75 & 0.29 & 0.91 & 0.92 &     0.76 \\
                Human F vs. Llama-3 (IM 2) & 0.97 & 0.75 & 0.58 & 0.65 & 0.76 & 0.56 & 0.28 & 0.81 & 0.89 &     0.69 \\
                Human F vs. Llama-3 (IM 3) & 0.97 & 0.87 & 0.84 & 0.92 & 0.96 & 0.90 & 0.58 & 0.91 & 0.93 &     0.88 \\ \midrule \midrule
                
    GPT-4o (IM 1) vs. Llama-3 (IM 1) & 0.99 & 0.62 & 0.80 & 0.66 & 0.89 & 0.85 & 0.91 & 0.56 & 0.95 &     0.80 \\
 GPT-4o (IM 1) vs. Gemini-1.5 (IM 1) & 0.82 & 0.89 & 1.00 & 0.96 & 0.99 & 0.99 & 0.77 & 0.89 & 0.93 &     0.92 \\
Llama-3 (IM 1) vs. Gemini-1.5 (IM 1) & 0.73 & 0.42 & 0.71 & 0.63 & 0.66 & 0.59 & 0.81 & 0.22 & 0.74 &     0.61 \\ [0.15cm]

 GPT-4o (IM 2) vs. Gemini-1.5 (IM 2) & 1.00 & 0.90 & 0.95 & 0.94 & 0.99 & 0.97 & 0.93 & 0.97 & 0.96 &     0.96 \\
    GPT-4o (IM 2) vs. Llama-3 (IM 2) & 0.99 & 0.72 & 0.87 & 0.67 & 0.90 & 0.71 & 0.84 & 0.71 & 0.91 &     0.81 \\
Gemini-1.5 (IM 2) vs. Llama-3 (IM 2) & 0.97 & 0.84 & 0.72 & 0.63 & 0.78 & 0.67 & 0.76 & 0.78 & 0.85 &     0.78 \\ [0.15cm]

    Llama-3 (IM 3) vs. GPT-4o (IM 3) & 0.99 & 0.98 & 0.91 & 1.00 & 0.97 & 1.00 & 0.99 & 0.99 & 1.00 &     0.98 \\
Llama-3 (IM 3) vs. Gemini-1.5 (IM 3) & 0.97 & 0.85 & 0.81 & 0.91 & 0.85 & 0.81 & 0.99 & 0.94 & 0.97 &     0.90 \\
 GPT-4o (IM 3) vs. Gemini-1.5 (IM 3) & 1.00 & 1.00 & 0.95 & 0.99 & 0.99 & 1.00 & 0.99 & 0.96 & 0.96 &     0.98 \\
\bottomrule
\end{tabular}
    
    \caption{Full AAR1 (adjacent agreement rate with $k=1$) results between all assessor pairs, evaluated at the level of each assessment criterion and the whole essay (``Overall''). C1: Material selection. C2: Material integration and citation; C3: Quality of key components. C4: Logic of structure. C5: Content and clarity of ideas. C6: Coherence (flow of ideas). C7: Cohesion (use of connectors). C8: Grammar and sentence structure. C9: Academic vocabulary.}
    \label{tab:AAR1_full}
\end{table*}

\subsection{Comments\label{app:comments}}
 
\begin{table*}[]
    \centering
    \small

\begin{tabular}{lllllllllll}
\toprule
Attr &          Assessor &             C1 &             C2 &              C3 &             C4 &             C5 &             C6 &              C7 &             C8 &             C9 \\
\midrule

  CR &               Human  B &            5.6 &           28.2 &             2.8 &            0.7 &            2.1 &            3.5 &            13.4 &           91.5 &           66.2 \\
   &                 Human C &          100 &          100 &           100 &           98.9 &          100 &           98.9 &           100 &          100 &          100 \\
   &                Human F &           99.1 &           97.2 &            96.2 &           80.2 &           87.7 &           83.0 &            89.6 &           90.6 &           82.1 \\ 
   &                 All LLMs &           100 &           100 &            100 &           100 &           100 &           100 &            100 &           100 &           100 \\ \midrule \midrule

    AL &                Human B &   26{\tiny±23} &   43{\tiny±32} &    59{\tiny±33} &    45{\tiny±0} &   50{\tiny±53} &   34{\tiny±26} &    46{\tiny±24} &  147{\tiny±83} &   97{\tiny±85} \\
   &                Human C &   17{\tiny±23} & 104{\tiny±122} &    39{\tiny±38} &   56{\tiny±77} & 112{\tiny±102} &   38{\tiny±69} &    26{\tiny±39} &  103{\tiny±88} &   65{\tiny±94} \\
   &                Human F &   26{\tiny±40} &   77{\tiny±75} &    51{\tiny±30} &   30{\tiny±39} &   51{\tiny±57} &   31{\tiny±31} &    27{\tiny±33} &   52{\tiny±67} &   79{\tiny±92} \\ [0.15cm]

  &     GPT-4o (IM 1) &   79{\tiny±10} &   82{\tiny±13} &     72{\tiny±8} &    59{\tiny±7} &   61{\tiny±10} &    53{\tiny±7} &     55{\tiny±9} &   59{\tiny±12} &   67{\tiny±10} \\
   & Gemini-1.5 (IM 1) &   98{\tiny±22} &  126{\tiny±29} &   120{\tiny±33} &   82{\tiny±19} &   91{\tiny±27} &   84{\tiny±22} &    85{\tiny±24} &   90{\tiny±26} &   99{\tiny±49} \\
   &    Llama-3 (IM 1) &   90{\tiny±13} &   91{\tiny±18} &    87{\tiny±14} &   60{\tiny±10} &   64{\tiny±15} &   54{\tiny±12} &    52{\tiny±15} &   56{\tiny±13} &   62{\tiny±13} \\ [0.15cm]

   &     GPT-4o (IM 2) &  291{\tiny±44} &  353{\tiny±40} &   332{\tiny±30} &  333{\tiny±37} &  374{\tiny±39} &  362{\tiny±42} &   347{\tiny±36} &  370{\tiny±45} &  357{\tiny±42} \\
   & Gemini-1.5 (IM 2) &  378{\tiny±88} & 446{\tiny±111} &  512{\tiny±106} & 399{\tiny±103} & 425{\tiny±121} & 397{\tiny±109} & 867{\tiny±2032} & 468{\tiny±148} & 400{\tiny±107} \\
   &    Llama-3 (IM 2) &  331{\tiny±35} &  368{\tiny±41} &  438{\tiny±111} & 466{\tiny±197} &  357{\tiny±68} & 351{\tiny±107} &   317{\tiny±88} &  345{\tiny±82} &  355{\tiny±97} \\ [0.15cm]

   &     GPT-4o (IM 3) &  295{\tiny±40} &  437{\tiny±40} &   372{\tiny±51} &  380{\tiny±40} &  444{\tiny±37} &  402{\tiny±39} &   358{\tiny±60} &  422{\tiny±50} &  321{\tiny±40} \\
   & Gemini-1.5 (IM 3) &  374{\tiny±86} & 654{\tiny±107} &  689{\tiny±110} &  592{\tiny±94} & 655{\tiny±102} & 559{\tiny±125} &  473{\tiny±109} & 642{\tiny±131} & 505{\tiny±343} \\ 
   &    Llama-3 (IM 3) &  333{\tiny±35} &  425{\tiny±53} &   409{\tiny±49} &  378{\tiny±51} &  481{\tiny±56} &  389{\tiny±51} &   366{\tiny±58} &  445{\tiny±62} &  362{\tiny±44} \\ \midrule \midrule

  PR &                Human B &             75 &            100 &             100 &            100 &            100 &             80 &              84 &            100 &             95 \\
   &                Human C &             19 &             72 &              54 &             50 &             81 &             40 &              29 &             79 &             80 \\
   &                Human F &             47 &             84 &              82 &             44 &             61 &             50 &              49 &             64 &             82 \\ 
   &                 All LLMs &           100 &           100 &            100 &           100 &           100 &           100 &            100 &           100 &           100 \\ \midrule \midrule
   
  AP &                Human B & 1.1{\tiny±1.0} & 2.1{\tiny±1.4} &  2.0{\tiny±1.2} &   1.0{\tiny±0} & 1.3{\tiny±0.6} & 1.0{\tiny±0.7} &  1.2{\tiny±0.9} & 5.3{\tiny±4.0} & 3.5{\tiny±3.0} \\
   &                Human C & 0.2{\tiny±0.6} & 2.1{\tiny±2.3} &  0.9{\tiny±1.1} & 1.1{\tiny±1.3} & 2.1{\tiny±1.8} & 0.9{\tiny±1.6} &  0.4{\tiny±0.6} & 2.5{\tiny±2.2} & 1.9{\tiny±1.8} \\
   &                Human F & 0.7{\tiny±1.0} & 2.4{\tiny±2.0} &  1.4{\tiny±1.0} & 0.8{\tiny±1.0} & 1.2{\tiny±1.4} & 0.8{\tiny±1.0} &  0.7{\tiny±0.9} & 1.4{\tiny±1.8} & 2.3{\tiny±2.2} \\ [0.15cm]
   
   &     GPT-4o (IM 1) & 1.8{\tiny±0.7} & 2.3{\tiny±0.8} &  3.4{\tiny±0.6} & 2.3{\tiny±0.8} & 2.0{\tiny±0.9} & 1.8{\tiny±0.7} &  1.3{\tiny±0.6} & 1.9{\tiny±0.7} & 2.2{\tiny±0.8} \\
   & Gemini-1.5 (IM 1) & 2.1{\tiny±0.8} & 2.6{\tiny±0.9} &  3.3{\tiny±1.0} & 1.9{\tiny±0.7} & 2.1{\tiny±0.8} & 2.5{\tiny±0.8} &  2.2{\tiny±0.7} & 2.4{\tiny±0.8} & 2.6{\tiny±1.5} \\
   &    Llama-3 (IM 1) & 2.2{\tiny±0.5} & 2.4{\tiny±0.6} &  3.1{\tiny±0.9} & 2.0{\tiny±0.7} & 2.3{\tiny±0.8} & 2.1{\tiny±0.6} &  1.5{\tiny±0.7} & 2.0{\tiny±0.7} & 2.3{\tiny±0.5} \\ [0.15cm]

   &     GPT-4o (IM 2) & 3.8{\tiny±0.8} & 4.8{\tiny±1.0} &  5.8{\tiny±1.5} & 4.6{\tiny±1.1} & 5.1{\tiny±0.9} & 5.5{\tiny±1.1} &  5.7{\tiny±1.2} & 5.0{\tiny±0.9} & 4.9{\tiny±1.1} \\
   & Gemini-1.5 (IM 2) & 5.0{\tiny±2.2} & 5.7{\tiny±2.5} &  8.2{\tiny±3.2} & 5.7{\tiny±2.6} & 6.1{\tiny±2.8} & 5.9{\tiny±2.7} &  5.7{\tiny±2.1} & 5.0{\tiny±2.2} & 5.4{\tiny±2.3} \\
   &    Llama-3 (IM 2) & 5.0{\tiny±1.7} & 5.7{\tiny±2.2} &  8.4{\tiny±3.0} & 8.1{\tiny±3.8} & 6.7{\tiny±2.7} & 6.9{\tiny±2.9} &  6.2{\tiny±1.9} & 6.1{\tiny±2.2} & 6.6{\tiny±2.4} \\ [0.15cm]
   
   &     GPT-4o (IM 3) & 3.9{\tiny±0.7} & 6.5{\tiny±1.7} &  8.5{\tiny±2.2} & 5.6{\tiny±1.1} & 7.7{\tiny±1.4} & 5.8{\tiny±1.0} &  5.1{\tiny±1.2} & 6.7{\tiny±1.8} & 5.2{\tiny±1.3} \\
   & Gemini-1.5 (IM 3) & 4.9{\tiny±2.2} & 9.0{\tiny±2.9} & 10.7{\tiny±3.4} & 9.3{\tiny±2.6} & 9.2{\tiny±2.8} & 8.2{\tiny±2.9} &  6.3{\tiny±1.6} & 9.1{\tiny±3.2} & 7.3{\tiny±3.4} \\
   &    Llama-3 (IM 3) & 5.1{\tiny±1.6} & 6.7{\tiny±2.2} &  8.9{\tiny±2.1} & 6.1{\tiny±2.1} & 7.6{\tiny±1.9} & 5.5{\tiny±1.8} &  5.4{\tiny±1.7} & 6.5{\tiny±2.0} & 5.9{\tiny±2.0} \\
\bottomrule
\end{tabular}

    \caption{General statistics of feedback comments generated by human and LLM assessors. CR (\%): comment rate, i.e., the percentage of time a comment is provided. AL: \textit{average length} (measured in tokens) of the provided comments (excluding cases where comments are not given), along with their respective standard deviations. PR (\%): problem rate, i.e., the percentage of time a problem is mentioned or implied in the \textit{provided comments}. AP: \textit{average number of problems} identified in the provided comments, along with their respective standard deviations. ``All LLMs'' means all three LLMs across the three interaction modes. C1: Material selection. C2: Material integration and citation; C3: Quality of key components. C4: Logic of structure. C5: Content and clarity of ideas. C6: Coherence (flow of ideas). C7: Cohesion (use of connectors). C8: Grammar and sentence structure. C9: Academic vocabulary.}
    \label{tab:generalStatsFeedbackComments}
\end{table*}

Table~\ref{tab:generalStatsFeedbackComments} presents the general statistics of feedback comments generated by human assessors and LLMs under the three interaction modes.

\subsection{Score-Comment Interaction\label{app:score-comment-interaction}}

\begin{figure*}
    \centering
    \includegraphics[width=1\linewidth]{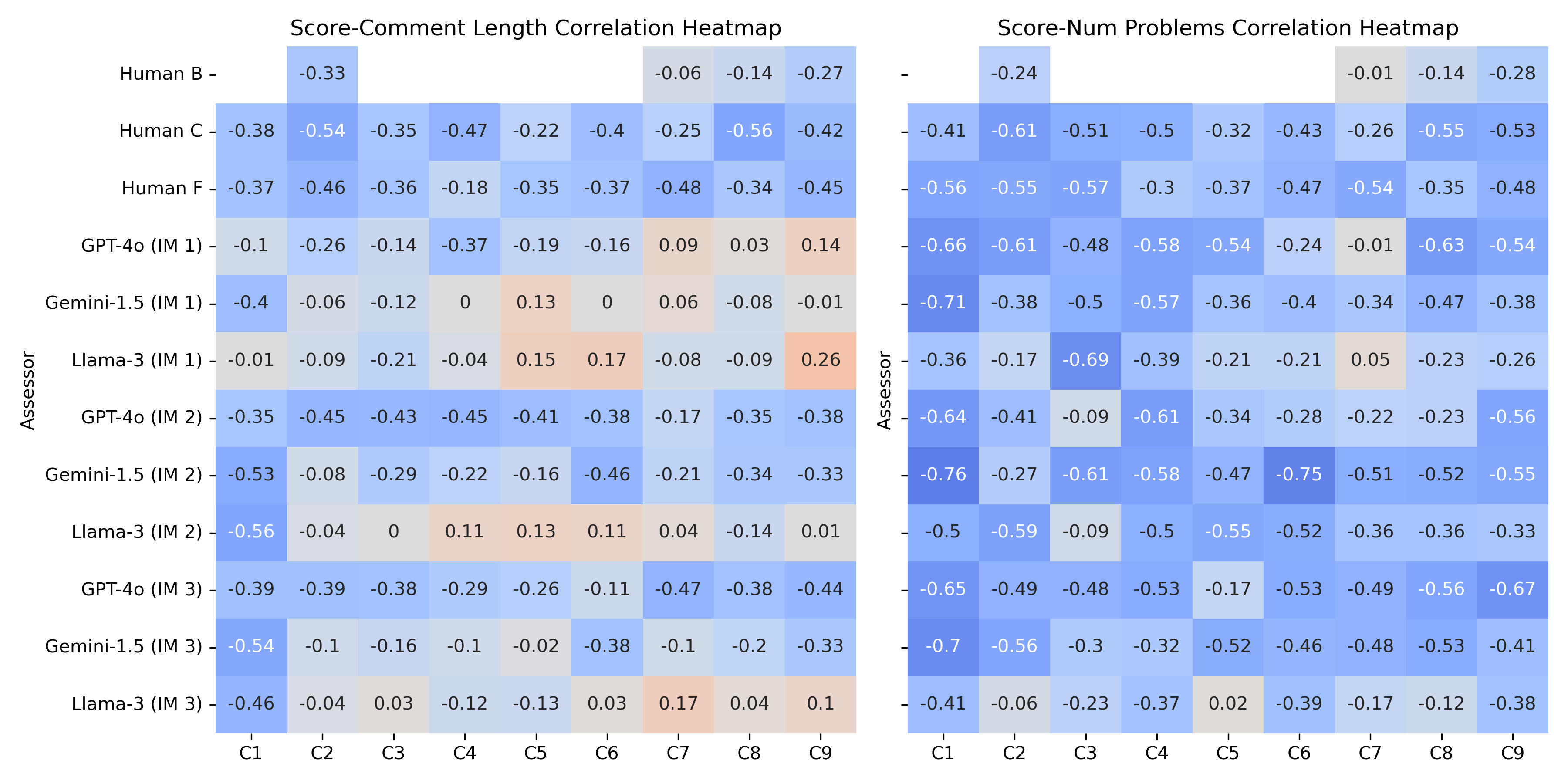}
    \caption{Heatmaps showing score-comment correlations between scores and the length of the related comments (left) and between scores and the number of problems identified in the related comments (right). Darker blue shades indicate a stronger negative correlation and darker orange shades a stronger positive correlation, with gray-ish colors indicating negligible correlations. To ensure meaningful analysis, correlations are calculated only when at least 10 score-comment pairs are available. C1: Material selection. C2: Material integration and citation; C3: Quality of key components. C4: Logic of structure. C5: Content and clarity of ideas. C6: Coherence (flow of ideas). C7: Cohesion (use of connectors). C8: Grammar and sentence structure. C9: Academic vocabulary.}
    \label{fig:score-comment-interaction}
\end{figure*}

Fig.~\ref{fig:score-comment-interaction} provides the full results of the correlations measured between scores and the token counts of or the numbers of identified problems in the related comments.

\section{Further Analyses\label{app:furtherAnalyses}}

Table~\ref{tab:commentPairs} provides five random example comment pairs sampled from GPT-4o-Aug and GPT-4o-May prompted under default prompt setting specified in Section~\ref{sec:prompting}. We find that when BERTScore is low (the last row), the comment pair is less similar compared to other pairs. While other two metrics (BLEU and ROUGE-L) are highly correlated with BERTScore (BLUE: 0.78, ROUGE-L: 0.88, Pearson), they consistently yield lower values than BERTScore. This indicates that these two lexical overlap-based metrics may be less effective at measuring comment reliability compared to the semantic similarity captured by BERTScore.

\begin{table*}[]
    \centering
    \scriptsize

\begin{tabular}{p{1cm}p{5cm}p{5cm}p{1cm}p{1.25cm}p{1cm}}
\toprule
                   Criterion &   GPT-4o-Aug &  GPT-4o-May &  BLEU &  ROUGE-L &  BERTScore \\
\midrule
  Content \& clarity of ideas &                                                    The content is generally relevant, but the clarity of ideas could be improved. Some sentences are complex and could be simplified for better understanding. For example, the sentence discussing Olson (2012) and Sanner (2008) could be broken down into shorter, more concise statements. Additionally, the review could benefit from more explicit explanations of key concepts, such as acculturation. &  The content is relevant and covers important aspects of the topic. However, some ideas are not clearly articulated. For example, the sentence "This is made worse because 'the nursing profession faces increased numbers of linguistically and culturally diverse patients who speak little or no English'" could be rephrased for clarity. Additionally, some sentences are overly complex and could be simplified for better understanding.&  0.20 &     0.30 &       0.63 \\ \midrule
          Logic of structure &                                                                                                                                                    The logical structure of the literature review could be improved. The review jumps between different topics without clear transitions, making it difficult for the reader to follow the argument. A more organized structure with clear subheadings and a logical flow of ideas would enhance readability. &                                                                                                                                                                                                                                                                                                                                                                                                         The logical structure of the literature review needs improvement. While the review covers important aspects of the psychological impact of COVID-19, the organization of the content is somewhat disjointed. Clearer subheadings and a more logical flow of ideas would help guide the reader through the review. For example, separating the discussion of different demographic groups or types of psychological impacts into distinct sections could enhance readability. &  0.17 &     0.41 &       0.70 \\ \midrule
          
   Quality of key components & The introduction provides a general overview of the topic but lacks a clear statement of the significance of the issue or specific research questions. The body of the literature review presents some relevant ideas but lacks a clear organization and focus on the pros and cons of online learning. The conclusion is missing, which is a significant omission as it should summarize key findings and highlight contributions or gaps in the literature. &  The introduction lacks a clear thesis statement and does not adequately set up the significance of the topic. The body of the review presents relevant ideas but lacks a clear structure and flow. The conclusion is weak and does not effectively summarize the key points or identify research trends or gaps. A more structured approach with clear sections and subheadings would improve readability and coherence. &  0.22 &     0.44 &       0.73 \\ \midrule
Grammar \& sentence structure &                                                                                                                                         The draft contains several grammatical errors and awkward sentence structures. For example, "The among of time is short considering the kind of assignments..." should be "The amount of time is short considering the type of assignments...". Varying sentence length and structure could also improve readability. &     The draft has several grammatical errors and issues with sentence structure. For example, "The among of time is short considering the kind of assignments" should be "The amount of time is short considering the type of assignments." Additionally, varying sentence length and structure would make the writing more engaging. Proofreading for grammatical accuracy and sentence variety is recommended. &  0.50 &     0.76 &       0.85 \\ \midrule
Grammar \& sentence structure &                                                                                                                                                                The grammatical accuracy is generally acceptable, but there are instances of awkward phrasing and run-on sentences. Varying sentence length and structure could improve readability. Additionally, ensuring subject-verb agreement and correct punctuation would enhance grammatical accuracy. & - Positive: The grammar is generally accurate.\newline- Improvement: Sentence length and variety need improvement. Some sentences are too long and complex, making them difficult to read.\newline- Example: "The health arguments against the use of cannabis include its addictive nature (Hurd et al., 2014) . It has also been directly linked to a range of adverse outcomes in physical health, which include lung cancer (Aldington et al., 2008), impaired respiratory function, cardiovascular disease, elevated systolic blood pressure, stroke (Singh et al., 2012), mental disorders (Saban et al., 2014), which include schizophrenia, especially amongst young people (Casadio et al., 2011), undesirable cognitive changes (Crean et al., 2011)." This could be broken down into shorter sentences. &  0.00 &     0.12 &       0.49 \\ 
\bottomrule
\end{tabular}

    \caption{Five random example comment pairs with their BLEU, ROUGE-L, and BERTScore scores provided. }
    \label{tab:commentPairs}
\end{table*}

\section{Prompts\label{app:prompts}}

Note that, any word followed by a dollar sign ``\$'' is a placeholder for all prompt templates included in this section. For example, ``\$comment'' is a placeholder for a comment.

\subsection{Prompts for the Feedback Comment Quality Evaluation Framework ProEval\label{app:promptFramework}}

The full prompt templates for the three steps in the pipeline of the feedback comment quality evaluation framework are given below. Among these three prompts, the prompt for Problem Extraction contains three in-context exemplars, whereas the prompts for the other two steps are zero-shot prompts. 

\subsubsection{Prompt for Problem Extraction}

\begin{quote}

\footnotesize

You will be given a feedback comment written for a student's essay. Your task is to identify and extract all the writing-related problems mentioned or implied in the comment, along with any explanations, suggestions, corrections, questions, quotations, or other relevant information provided in the comment for each extracted problem. \newline

A writing-related problem is any issue that affects the quality of the writing, such as citation errors,
logical flaws, coherence issues, grammatical mistakes, or inappropriate word choices, among others. \newline

\#\#\# Extraction Instructions\newline

- Each extracted problem must be clear and can be understood without the need to refer to the original comment. \newline

- Each extracted problem must faithfully reflect the provided comment by including any relevant information. Relevant information includes a further explanation or an elaboration of the problem, a suggestion for improvement, a concrete correction, a clarifying question, an excerpt (possibly without quotation marks) from the student's essay, or any other relevant information that helps to understand the problem. \newline

- Whenever possible, extract each problem and the relevant information as they are written in the comment. \newline

\#\#\# Output Instructions\newline

- Output each extracted problem along with their relevant information line by line headed by ``-''.
- Output ``None'' if no writing-related problems are mentioned or implied in the comment.\newline

\#\#\# Examples\newline

Example 1 input:\newline

The content is generally informative and relevant, but the clarity of ideas could be improved. Some sentences are overly complex and could be simplified for better understanding. For instance, the sentence ``Gandhi's Satyagraha as an adequate substitute for violent methods of conducting social conflict in an early and thorough philosophical examination of Gandhi's attitude to violence in extreme group conflict'' is difficult to parse and could be rephrased for clarity.\newline

Example 1 output: \newline

- The clarity of ideas could be improved. Some sentences are overly complex and could be simplified for better understanding. For instance, the sentence ``Gandhi's Satyagraha as an adequate substitute for violent methods of conducting social conflict in an early and thorough philosophical examination of Gandhi's attitude to violence in extreme group conflict'' is difficult to parse and could be rephrased for clarity.\newline

Example 2 input:\newline

The content and clarity of ideas are generally good, but there are some areas where the author could provide more depth or analysis. For example, the author could have explored the potential reasons why students in India may be more vulnerable to substance abuse, or discussed the implications of legalization for public health policy. To improve, the author could revisit the body of the literature review and provide more nuanced analysis of the findings.\newline

Example 2 output:\newline

- There are some areas where the author could provide more depth or analysis. For example, the author could have explored the potential reasons why students in India may be more vulnerable to substance abuse, or discussed the implications of legalization for public health policy. To improve, the author could revisit the body of the literature review and provide more nuanced analysis of the findings.\newline

Example 3 input:\newline

The author has generally done a good job of integrating source materials and presenting information clearly. However, there are some instances where the connections between ideas could be more explicitly stated, and the citation practices could be more consistent (e.g., some sources are cited with author names, while others are cited with only the year).\newline

Example 3 output:\newline

- There are some instances where the connections between ideas could be more explicitly stated.\newline
- The citation practices could be more consistent (e.g., some sources are cited with author names, while others are cited with only the year).\newline

\#\#\# Input\newline

\$comment\newline

\#\#\# Output

\end{quote}

\subsubsection{Prompt for Problem Classification}

\begin{quote}
\footnotesize

You will be given an excerpt of a feedback comment written for a student's essay. Your task is to answer the following questions: \newline

1. Does the excerpt refer to a specific part of the essay? A specific part refers to a part of the essay that can be easily located by the student.
For example, it can be a specific word, phrase, sentence, paragraph, reference etc. used in the essay. It can be a concrete location, such as ``sentence 2 in paragraph 2,'' ``in paragraph 6,'' ``the first citation,'' or ``the first sentence of the paper'' and so on. A less concrete location, such as ``the introduction,'' or ``the conclusion,'' is also considered a specific part if it is accompanied by some referenceable details, such as ``The significance of South Australian policy is unclear, as it is the first citation and the only one in the Introduction.'' Note that the excerpt may only contain a quoted text from the essay, in which case, the quoted text is considered a specific part.\newline

2. Does the excerpt offer some form of suggestions, general or specific, for the student to improve the essay? If the excerpt only describes a problem and it is unclear what the student should do to fix it, then there is no suggestion. If the excerpt provides a concrete correction, it is considered a suggestion.\newline

3. Does the excerpt provide a concrete correction for the student to apply? Note that when the excerpt only contains a quoted text from the essay and there are some notes indicating a correction (e.g., adding/removing a punctuation, correcting a spelling), this is considered a correction.\newline

Answer each question with ``Yes'' or ``No'' based on the content of the excerpt and briefly justify your answer. After answering all the questions, 
produce your final answers in a newline separated by commas.

Excerpt: \$excerpt

\end{quote}

\subsubsection{Prompt for Correction Relevancy Check}

\begin{quote}

\footnotesize

You will be given an excerpt of a feedback comment written for a student's essay according to an assessment question. Your task is to answer the following questions: \newline

1. Does the problem pointed out in the excerpt exist in the corresponding essay? If the excerpt uses a quoted text to point out a problem, check if the quoted text is present in the essay. Please note that the quoted text may not be an exact match either due to misspellings, capitalization errors etc., or because the quoted already contains the correction in place. \newline

2. Is the problem pointed out in the excerpt relevant to the corresponding assessment question? Check if the excerpt is broadly related to any aspect of the assessment question. \newline

3. Is the correction of the problem pointed out in the excerpt correct? If the problem does exist in the essay, check if the correction fixes the problem or presents a plausible solution or improvement. \newline

Here is the essay: \newline

\$essay \newline

Here is the assessment question:\newline

\$question \newline

Here is the excerpt:\newline

\$excerpt\newline

Answer each question with ``Yes'' or ``No'' utilizing all the information provided and briefly justify your answer. After answering all the questions, produce your final answers in a newline separated by commas.
    
\end{quote}

\subsection{Prompts for the Main Experiments\label{app:mainPrompts}}

Our prompts consist of three parts: (1) a system prompt part that provides general background information and specifies the writing topic and some general assessment guidance; (2) a writing part that includes an entire literature review (with references); (3) an assessment instruction part, where one or multiple assessment questions (see Table~\ref{tab:assessmentCriteria}) are asked in various manners according to the interaction modes.  

We keep the system prompt fixed across the three interaction modes. For the main experiments, the system prompt is as follows:

\begin{quote}
\footnotesize

You are an expert academic writing instructor specializing in graduate-level work, with particular experience supporting students who speak English as an additional language. You have been asked to evaluate a literature review submitted by a graduate student on the following topic: \$Topic. The review was written in 2021, so references after this year are not expected.

When assessing the student's writing, please strictly follow the instruction provided to you and make sure your score/feedback is carefully considered and constructive. Please provide your comments and/or suggestions with as much detail and specificity as possible. Please provide specific examples of sentences, paragraphs or sections that you think could use improvement. If you write comments, please start them with something positive. Please proceed with things that could be improved, would make things clearer for the reader, would make the text flow better, etc.

\end{quote}

For the writing part, we explicitly mark the beginning and end of the writing for clarity: 

\begin{quote}
\footnotesize

\#\#\#\#\#\#\#\#\#\# Writing starts \#\#\#\#\#\#\#\#\#\#

\$writing

\#\#\#\#\#\#\#\#\#\# Writing ends \#\#\#\#\#\#\#\#\#\#
\end{quote}

The specifics of how the assessment instruction part is constructed are detailed below.

\subsubsection{Interaction Mode 1}

In Interaction Mode 1, all assessment questions (see Table\ref{tab:assessmentCriteria}) are asked at once:

\begin{quote}
\footnotesize

Q1: \{Assessment question 1\}

Q2: \{Assessment question 2\}

...

Q9: \{Assessment question 9\}

\end{quote}

After these assessment questions is an answer instruction: 

\begin{quote}
\footnotesize

For each of the 9 questions above, provide your comments or suggestions if any, followed by your score out of 10. Please indicate which question you are providing feedback for by starting your response with `A1:', `A2:', etc. Each response should use the following format:

Score: ...

Comments or suggestions: ...

\end{quote}

Note that we use ``if any'' to denote the optionality of the comments and suggestions. We tried putting ``(Optional)'' after ``Comments or suggestions,'' but that does not make a difference.

\subsubsection{Interaction Mode 2}

In Interaction Mode 2, the assessment questions are presented sequentially and one at a time. Below is the basic structure:

\begin{quote}
\footnotesize

Q$_i$: \{The $i$th assessment question.\}

\{Answer instruction\}

A$_i$:
    
\end{quote}

The answer instruction resembles the one used in the Interaction Mode 1. 

\begin{quote}
\footnotesize

Provide your score out of 10, followed by comments or suggestions if any. Your response should use the following format:

Score: ...

Comments or suggestions: ...
    
\end{quote}

Note that, we append LLM's response to the $i$th assessment question to the original prompt to form a new prompt, to which the next assessment question is added. This way, the writing is only provided once (at the beginning), but the LLM will have access to previous assessment questions as well as its answers to those questions.

\subsubsection{Interaction Mode 3}

In Interaction Mode 3, each assessment question is asked independently, so there are 9 separate prompts for each essay. 

The structure for the assessment part of the prompt is similar to that in Interaction Mode 2, but without indexation and prefix ``Q/A'':

\begin{quote}
\footnotesize

\{An assessment question.\}

\{Answer instruction\}
    
\end{quote}

The answer instruction works exactly the same as in Interaction Mode 2.

\subsection{Prompts for the Follow-Up Experiments\label{app:followUpPrompts}}

\subsubsection{System Prompt Simplification}

Below is a simplified system prompt removing the helpful information from the default system prompt used in Section~\ref{sec:experiments}.

\begin{quote}
    \footnotesize
You are an expert academic writing instructor for graduate students. You have been asked to evaluate a literature review submitted by a student below. The writing is broadly related to the following topic: \$Topic. 

When assessing the student's writing, please strictly follow the instruction provided to you and make sure your score/feedback is carefully considered and constructive.
\end{quote}

\subsection{Prompts for Assessing Specificity and Helpfulness}

\begin{quote}
    \footnotesize

You will be given a feedback comment written for a student’s essay according to an assessment question. Your task is to rate the feedback comment on (1) specificity and (2) helpfulness, using a scale from 1 to 10, where 1 is the lowest and 10 is the highest. Conclude your response with the final ratings in this format: "Specificity: X, Helpfulness: X" (where X is a score from 1 to 10).

Here is the essay:

\$essay

Here is the assessment question:

\$question

Here is the feedback comment:

\$feedback

Please rate the specificity and helpfulness of the feedback comment.

\end{quote}

\end{document}